\documentclass[preprint,12pt,3p]{elsarticle}

\usepackage{amssymb}
\usepackage{amsmath}
\usepackage{url}
\usepackage{caption}
\usepackage{subcaption}
\usepackage{color}

\newcommand{\df}[1]{\textcolor{black}{ #1}}
\newcommand{\mn}[1]{\textcolor{black}{ #1}}

\journal{Computer Speech and Language}

\begin{document}

\begin{frontmatter}

\title{DNN Adaptation by  Automatic Quality Estimation \\ of ASR Hypotheses}






\author[label1]{Daniele Falavigna}
\ead{falavi@fbk.eu}

\author[label1]{Marco Matassoni\corref{cor1}}
\address[label1]{FBK-Fondazione Bruno Kessler,  Trento, Italy}
\address[label2]{University of Trento, Trento, Italy}

\cortext[cor1]{\textcolor{red}{Please cite this article as: D. Falavigna et al., DNN adaptation by automatic quality estimation of ASR hypotheses,
Computer Speech \& Language (2016), http://dx.doi.org/10.1016/j.csl.2016.11.002}}

\ead{matasso@fbk.eu}

\author[label1,label2]{Shahab Jalalvand}
\ead{jalalvand@fbk.eu}

\author[label1]{Matteo Negri}
\ead{negri@fbk.eu}

\author[label1]{Marco Turchi}
\ead{turchi@fbk.eu}

\begin{abstract}
In this paper we propose to exploit the automatic Quality Estimation (QE) of ASR hypotheses to perform the unsupervised adaptation of a deep neural network modeling acoustic probabilities. Our hypothesis is that significant improvements can be achieved by: \textit{i)} automatically transcribing  the evaluation data we are currently trying to recognise, and \textit{ii)} selecting from it a subset of ``good quality'' instances based on the word error rate (WER) scores predicted by a QE component. To validate this hypothesis, we run several experiments on the evaluation data sets released for the CHiME-3 challenge. First, we 
operate in oracle conditions in which manual transcriptions of the evaluation data are available, thus allowing us to compute the \textit{true} sentence WER. In this scenario, we perform the adaptation  with variable amounts of data, which are characterised by different levels of quality. Then, we move to realistic conditions in which the manual transcriptions of the evaluation data are not available. In this case,
the adaptation is performed on data selected according to the WER scores \textit{predicted} by a QE component. Our results indicate that: \textit{i)} QE predictions allow us to closely approximate  the adaptation results obtained in oracle conditions, and \textit{ii)} the overall ASR performance based on the proposed QE-driven adaptation method is significantly better than the strong, most recent, CHiME-3 baseline.
\end{abstract}


\end{frontmatter}


\section{Introduction}
\label{sec:intro}

Automatic speech recognition (ASR) with microphone arrays is gaining increasing interest in a variety of application scenarios, such as home and office automation, smart cars and  humanoid robots. In such applications, ASR should be able to operate in environments where noises of various types, competing speakers and reverberation  heavily affect recognition performance, which is usually satisfactory in controlled acoustic conditions.
To cope with the above scenarios, most of the current approaches are based on the implementation of a variety of enhancement techniques such as beamforming, denoising and dereverberation~\cite{brandstein2001}.

The last CHiME challenge (CHiME-3\footnote{\url{http://spandh.dcs.shef.ac.uk/chime_challenge/}}) provided an excellent framework to evaluate signal enhancement approaches and noise-robust acoustic modelling techniques for ASR. Participants' results  \cite{CHiME3}  evidenced the effectiveness of signal enhancement approaches, mostly  based on ``beamforming'', combined with the use of hybrid acoustic models based on  deep neural networks hidden Markov models (DNN-HMMs) \cite{hinton2012,mohamed2012,renals2013,renals2014}. The effectiveness of acoustic modelling based on context-dependent  DNN-HMMs  was also demonstrated in several works dealing with applications spanning from mobile voice search \cite{dahl2012} to the transcription of broadcast news and YouTube videos \cite{hinton2012}, conversational (at the telephone or in live scenarios) speech recognition \cite{telseide2011} and ASR in noisy environments \cite{seltzer2013}.

In \cite{falavi2015}, we observed a significant WER reduction  on the CHiME-3 \textit{real} test data by retraining the baseline DNN on the evaluation set itself and  using the automatic transcriptions resulting from a first decoding pass to align acoustic observations with DNN outputs. After this ``unsupervised'' DNN retraining step (unsupervised as it relies on automatic transcriptions, which are not revised by a human) we achieved a WER reduction from 20.2\% to 15.5\%.\footnote{This performance was achieved using filter-bank log-energies as acoustic features, as included in a previous CHiME-3 ASR baseline used in the competition held  in 2015 (\url{http://spandh.dcs.shef.ac.uk/chime_challenge/}). 
A stronger baseline, which employs speaker normalized features, has been successively delivered. This baseline is the one used in the experiments documented in this paper.}
These positive results motivate the research proposed in this work, which further explores unsupervised techniques for the ``self'' adaptation of a DNN (i.e. by tuning its parameters on the same test set we are currently trying to recognize) with an improved, automatically generated supervision. In particular, the full DNN retraining step adopted in \cite{falavi2015} is now substituted by a more sophisticated solution, which enhances the adaptation with effective instance weighing and selection criteria.

At its core, our adaptation method is similar to the one described  in \cite{yu2013}, which adds to the objective function  to optimize a regularization term based on the Kullback-Leibler divergence (KLD) between the original (non adapted) and the current DNN output distribution.\footnote{The reason for choosing KLD regularization is that it can be implemented by constraining directly the target output DNN probabilities,  making it possible to easily integrate the approach in the
existing open-source ASR toolkits like KALDI \cite{Povey_ASRU2011}.}
However, departing from Yu et al.'s method, we explore different ways to enhance the process with automatic ASR quality predictions. In particular, building on the outcomes of previous research  on automatic quality estimation  for  ASR \cite{Negri:2014,sjalalvand-EtAl:2015:ACL}, we focus on two alternative solutions. The first one is based on \textit{weighing} the  KLD regularization term  with coefficients that depend on the predicted quality of each transcribed sentence. The second one is based on \textit{filtering} the adaptation set by removing the utterances that, in terms of predicted quality, seem to be less reliable.

Our evaluation is carried out with the data sets released for the  CHiME-3 challenge and  with the most recent CHiME-3 baseline system. In a first round of experiments, we adapt the original DNN in {\em cross conditions}. In this setting,  the manually transcribed development set is distinct from the evaluation set and it is used for a supervised DNN adaptation process. The available supervision allows us to compare the
resulting performance with the results achieved when automatic  (hence less accurate) 
transcriptions are used in place of the manual ones.  With an oracle-based sentence selection (obtained by using the manual transcriptions as references to calculate the true sentence WER) we observe significant performance improvements when  subsets of good quality adaptation instances are used instead of the whole data. This finding supports the intuition that automatic techniques to estimate transcription quality (e.g. by filtering out those with higher WER) can be used to inform the adaptation process and increase its robustness.

In the second round of experiments we switch to self DNN adaptation in {\em homogeneous conditions}, in which  the adaptation is performed with automatic transcriptions of the same evaluation data we are currently trying to recognise (i.e. the \textit{real} test set used in the CHiME-3 challenge). In this scenario we achieve the most interesting results, with a relative WER reduction of 11.7\% (from 15.4\% to 13.6\%) obtained with the automatic sentence-level WER predictions returned by our ASR QE component. Besides this, another interesting result of our experiments is that  the  ASR performance gain achieved through self DNN adaptation has to be mostly attributed to  the automatic selection of the adaptation utterances rather than to weighing the KLD regularization term, which has a limited effect on the overall performance.
To complete our analysis, the usefulness of the proposed QE-based adaptation method is verified not only with filter-bank features, but also with feature normalization via maximum likelihood linear regression (fMLLR) transformations, which characterize the best performing systems in the CHiME-3 challenge  \cite{hori2015,yoshioka2015} as well as the most recent baseline. This is an interesting result since, while DNN adaptation  has already proven to be effective with filter-bank features (see Section \ref{sec:related} for a review of DNN adaptation approaches), the benefits yielded by adaptation using  acoustic features (speaker) normalized through fMLLR transformations are still questionable.

To the best of our knowledge, this paper represents the first investigation of the use of QE-based sentence selection  for unsupervised DNN adaptation in the framework of ASR decoding. Overall, its main contributions include:

\begin{itemize}
    \item A new application of the ASR QE procedure described in \cite{Negri:2014} to predict the WER of automatic transcription hypotheses;
    \item An extension of the KLD regularization approach for unsupervised DNN adaptation \cite{yu2013}, which could be easily integrated in the KALDI speech recognition toolkit \cite{Povey_ASRU2011}; 
    \item Significant improvements over the strong, most recent CHiME-3 baseline. 
\end{itemize}

All the experiments described in this paper have been carried out with the TranscRater open-source tool described in \cite{jalalvand-EtAl:2016:P16-4}.

The paper is organized as follows. 
In Section~\ref{sec:related} we summarize relevant previous works related to our research. 
In Section~\ref{sec:adaptation} we describe our approach to  DNN adaptation. 
In Sections~\ref{sec:quality} and \ref{sec:ASR} we 
present the adopted automatic WER prediction method and the ASR system architectures.
After the description of our experimental setup in Section~\ref{sec:setup}, our results are presented in Section~\ref{sec:results} and discussed in
Section~\ref{sec:res_fmllr}. 

\section{Related work}
\label{sec:related}

Detailed overviews about the CHiME challenges of years 2011 (CHiME-1), 2013 (CHiME-2) and 2015 (CHiME-3) can be respectively found in \cite{CHiME1,CHiME2,CHiME3}. While the first round in 2011 was mostly focused on speech separation task, the last two editions addressed large vocabulary speech recognition in noisy environments. The last one, in particular, involved many participants addressing a variety of topics such as noise reduction, de-reverberation, speaker/noise adaptation, system combination and rescoring with long-spanning LMs. Our submission \cite{falavi2015} mostly focused on three aspects:
 \textit{i)} the automatic selection of the best channel,  \textit{ii)} DNN retraining and \textit{iii)} rescoring of word lattices with a linear combination of 4-grams LMs and RNNLMs. As mentioned in the introduction, the significant improvements achieved with unsupervised DNN retraining motivated us to further investigate the DNN adaptation issue.

In the past, several adaptation techniques have been proposed for artificial neural networks employed in ASR hybrid systems. These are mostly based on the estimation  of linear transformations 
of their input, output or hidden units \cite{gemello2007,abrash95,neto95,li2010,siniscalchi2013}. Feature discriminative linear regression (fDLR) \cite{seide2011} and output-features discriminative linear regression (oDLR) \cite{yao2012} are other approaches specifically investigated for DNN adaptation.
Regardless of the layer to which the transformation is applied, in all the mentioned approaches only the weights of the linear transformation are updated in order to optimize an objective function computed on the adaptation data. In this way, the probability that the DNN model overfits the adaptation data is reduced. Note that both fMLLR and fDLR are linear transformations applied to the input features; the difference between the two lies in the  estimation criterion they adopt. fMLLR maximizes the likelihood of the adaptation observations, while fDLR optimizes a discriminative criterion computed on the same adaptation observation (e.g. it minimizes the mean squared error between target and actual output-state network distribution). 

A variant of fDLR is described in 
\cite{huang2014}, which  proposes to adapt the DNN parameters within  a maximum a posterior (MAP) framework. Basically, the method  consists in adding to the objective function to optimize  a term representing the prior density of the 
linear transformation weights. This approach demonstrated to be equivalent to L2 norm regularization \cite{bilmes2006} if the prior distribution of transformation weights is assumed to be Gaussian ${\cal{N}}(0,I)$. In general, adding a regularization term to the objective function proved to be effective to reduce model overfitting.
An excellent review of ``conservative training'' approaches for artificial neural networks can be found in \cite{laface2006}. The use of a momentum term to update the DNN weights, the use of small values for the learning rate as well as of an early stopping  criterion can be considered as adaptation methods.

In \cite{yu2013}, the Kullback-Leibler divergence (KLD) between the original unadapted distribution of the DNN outputs and the related distribution estimated on the adaptation set is considered as  regularization term. As reported in \cite{yu2013} 
this approach, also employed 
in our work,  allowed obtaining significant WER reductions compared to fDLR transformation of the input features on two different tasks: voice search and lecture transcription. 
The weight assigned to the regularization term in the objective function is an important parameter to choose when using regularized learning.

The use of fMLLR  features in combination with hybrid DNN-HMMs has been studied in \cite{amazon2015}. On a private clean speech evaluation set, the authors observed that:
{\em i)} filter-bank features and fMLLR  features achieved comparable performance, and  {\em ii)} only the combination of the two types of features, either at an early or late fusion stage,
provided significant WER reductions. These results are somehow in contrast with those obtained in the CHiME-3 challenge, in which
fMLLR normalization gives significant improvements compared to speaker-independent filter-bank features. However, we have to consider that participants in the CHiME-3 challenge experimented on noisy data  and did not apply any automatic speaker diarization module, since both utterance segmentation and speaker labels were manually checked.

\df{An approach for  unsupervised speaker adaptation of DNNs using fMLLR features is also reported in  \cite{Swiet2014}. The authors propose to train 
speaker-dependent amplitude parameters associated to hidden units of the network, obtaining significant performance improvements on the recognition of English TED talks.}

\df{In the context of speaker-adaptive training (SAT) via fMLLR \cite{gales98}, recently proposed  approaches  make use of i-vector \cite{kenny2008} as speaker representation to perform acoustic feature normalization. In \cite{miao2015} an adaptation neural network is trained to convert i-vectors to speaker-dependent linear shifts which, in turn, are used to generate speaker-normalized features for SAT-DNN training/decoding. The work reported in  \cite{garimella2015} proposes to process HMM-based i-vectors with specific hidden layers of a DNN before  combining them with hidden layers  processing standard acoustic features. The work reported in  \cite{Kara2015} proposes to incorporate prior statistics (derived from gender clustering of training data) into i-vectors estimation, showing significant perfomance improvements when the approach is used for DNN adaptation of a hybrid ASR system.}

The automatic selection of training data for acoustic modelling in speech recognition has been previously addressed  in the context of lightly supervised training \cite{Lamel2001} and active learning approaches \cite{riccardi2002,falavignaSC2006}. 
The use of confidence measures for improving MLLR transformations has also been investigated by \citep{Pitz00improvedmllr} in 
a German conversational speech recognition task. The authors 
showed significant WER reductions, for an ASR system based on a Gaussian Mixture Model (GMM),  by removing the low confidence frames from the adaptation data. 
More recently, \cite{thomas2013}  proposed an automatic sentence selection method based on different types of confidence measures for the semi-supervised training of DNNs in a low-resource setting.

The use of QE as a quality prediction method alternative to confidence estimation is inspired by  previous research on QE for machine translation \cite{Mehdad2012a,camargodesouza-EtAl:2013:WMT,turchi-negri-federico:2013:WMT,Souza2014}. In the ASR field it
has been first proposed in \cite{Negri:2014}.
In such previous work, \df{the} objective was to  bypass the dependency of confidence estimation on knowledge about the inner workings of the decoder that produces the transcriptions and, in turn, to avoid the risk of 
biased (often overestimated) quality estimates \cite{Woodland2000}. 
In \cite{Negri:2014} ASR QE is explored as a supervised regression problem in which the WER of an utterance transcription  has to be automatically predicted.  
The extensive experiments in different testing conditions discussed in \cite{Negri:2014} indicate that  regression models based on Extremely Randomized Trees (XRT)
\cite{geurts2006extremely}
can achieve competitive performance, being able to outperform strong baselines and to approximate the true WER scores computed against reference transcripts.
In \cite{cdesouza-EtAl:2015:NAACL-HLT}, our basic approach was refined in order to achieve robustness to large differences between training and test data. The proposed domain-adaptive approach based on multitask learning was intrinsically evaluated on multi-domain data, achieving good results both in regression and in classification mode.
In order to explore the possible applications of ASR QE, in \cite{sjalalvand-EtAl:2015:ACL} we proposed its use for successfully improving hypothesis combination with  ROVER \cite{fiscus1997post}.
Finally, in \cite{jalalvand-EtAl:2016:P16-4}, we described TranscRater, our recently released open-source ASR QE tool. The tool is the one used for the experiments described in this paper.


\section{KL-divergence based regularization}
\label{sec:adaptation}

The DNNs considered in this work estimate the posterior probability of an output unit $s_i$ associated to a HMM output probability density function (PDF). The state posterior probability $p[s_i|o_t]$, being $o_t$ an observation at time $t$,  is then converted into a PDF using the following Bayes formula:

\begin{equation}
p[o_t|s_i] = p[s_i|o_t]\frac{p[o_t]}{p[s_i]} \ \ \ 1\leq i\leq I
\end{equation}

where $I$ is the total number of output PDFs and $p[o_t]$ is discarded since it does not depend on the state.

A possible criterion for estimating weights and biases of the DNN is to minimize over a training sample  the cross-entropy $\mathcal{C}(\hat{p},p)$ between a target distribution $\hat{p}$ and the estimated one:

\begin{equation}
\mathcal{C}(\hat{p},p)=\frac{1}{T}\sum_{t=1}^{T}\sum_{i=1}^{I}\hat{p}[s_i|o_t]\log p[s_i|o_t]
\end{equation}

where $T$ is the total number of frames in the training utterances. Usually, the entries $\hat{p}[s_i|o_t]$ in the target distribution are obtained by forced alignment using an existing ASR system and assume the value of $1$ over the aligned states.

The usual way to adapt a DNN trained on a large set of data (e.g. some thousands of hours of speech), given a much smaller  set of adaptation data (e.g. some minutes of speech\footnote{Uttered by a new speaker or recorded in a new acoustic environment.}), is to retrain the DNN over the adaptation set. 
This approach assumes that either manual or automatic transcriptions of the adaptation sentences are available but, due to the small size of the adaptation set compared with the high number of parameters of the DNN model,
it could happen that the model overfits the training data. As mentioned in the previous section, a solution to prevent overfitting effects is to adopt a conservative learning procedure, which can be implemented by adding a regularization component to the loss function to optimize.

In \cite{yu2013} the authors propose to use   the Kullback-Leibler divergence between the original distribution  and the adapted  one as regularization term. Adding KLD computed on the adaptation set to cross-entropy (also evaluated on the adaptation set)
results in the following objective function to minimize:
\begin{small}
\begin{equation}
\mathcal{D}(\overset{*}{p},p)=(1-\alpha)\mathcal{C}(\hat{p},p)+\alpha \frac{1}{N}\sum_{t=1}^{N}\sum_{i=1}^{I}\overset{*}{p}[s_i|o_t]\log p[s_i|o_t]
 \label{eq:div0}
\end{equation}
\end{small}
where $N$ is the number of adaptation frames, $\overset{*}{p}[s_i|o_t]$ is the posterior probability computed with the original DNN and $\alpha$ is the regularization coefficient. As reported in \cite{yu2013}, Equation~\ref{eq:div0} can be rewritten as follows:
\begin{equation}
\mathcal{D}(\overset{*}{p},p)=\frac{1}{N}\sum_{t=1}^{N}\sum_{i=1}^{I}P[s_i|o_t]\log p[s_i|o_t]
\label{eq:div1}
\end{equation}
where
\begin{equation}
P[s_i|o_t]=(1-\alpha)\hat{p}[s_i|o_t]+\alpha \overset{*}{p}[s_i|o_t] \ \ \ \  0\leq\alpha\leq 1
\label{eq:newp}
\end{equation}
 
Equation~\ref{eq:div1} states that KLD regularization can be implemented through cross-entropy minimization between a new target probability distribution $P$ and the current probability distribution $p$. The new target  distribution is obtained as a linear interpolation  of the original distribution $\overset{*}{p}$ and the distribution $\hat{p}$ computed via forced alignment with the adaptation data. Note that, in Equation~\ref{eq:newp}, a value of $\alpha=0$  is equivalent to do a ``pure'' retraining of the DNN over the adaptation data (i.e. completely trusting them), while a value of $\alpha=1$ means that the output probability distribution of the  adapted DNN is forced to follow that of the  original DNN (i.e. completely trusting the original model). Usually, the value of $\alpha$ is estimated on a development set, together with the value of the learning rate, and does not change across the test utterances. What one can expect is that the optimal value of $\alpha$ is close to 0 when the size of the adaptation set is large and the transcriptions of the adaptation sentences are not affected by errors (i.e. in supervised conditions). 
Otherwise, when the size of the adaptation set is small and/or its transcription can be affected by errors (i.e. in the case of unsupervised adaptation), the optimal value of $\alpha$ should increase. 

It is worth remarking that the original DNN, producing the distribution $\overset{*}{p}$ used in the above equations, could have been  trained by optimizing a criterion different from cross-entropy minimization. This is actually the approach used in this study (see Section~\ref{sec:ASR} for details on baseline DNN training).

Finally, unlike other methods,  note that  KLD-based regularization  binds directly the DNN output probabilities rather than the model parameters. In this way, the method can be easily implemented with any software tool based on back-propagation \mn{(e.g. the KALDI toolkit)}, without introducing any modification.

\subsection{Soft DNN adaptation}
\label{subsec:soft-ada}

Experiments in \cite{yu2013} have shown a dependency of the optimal value of $\alpha$ in Equation~\ref{eq:newp} on the size of the adaptation data. However, as mentioned above, one could also expect that the optimal value of $\alpha$ depends on the quality of the supervision. 
%
Starting from this intuition, here
we propose to compute $\alpha$  on a sentence basis, as a function of   sentence WER estimates. To this end, 
we take advantage of previous research we conducted on ASR quality estimation (WER prediction) and word error detection.


\df{In principle, we could simply use as sentence-dependent regularization coefficient the following value: $\alpha(k)=\mathrm{WER}^{pred}_k, 1\leq k\leq K$, where $0\leq\mathrm{WER}^{pred}_k\leq1$ is an automatic estimate of the $k^{th}$ sentence WER  and $K$ is the total number of adaptation sentences. However, note that in doing this if the value of $K$ is small and $\mathrm{WER}^{pred}_k\cong 0, \forall k$ the original distribution $\overset{*}{p}$, in Equation~\ref{eq:newp}, is weighted by $\alpha\cong 0$ (i.e. we completely trust the adaptation data), augmenting the risk that the adapted  DNN overfits the adaptation data.
To avoid this effect we can simply  add a bias to the sentence WER estimate as follows:}

\begin{equation}
\alpha(k)=\beta+(1-\beta)\times \mathrm{WER}^{pred}_k  \ \ \ \ \ \ \ \  1\leq k\leq K \ \ \ \ \ 0\leq \beta\leq 1
\label{eq:alfasent}
\end{equation}
In the equation above, a value of $\beta=0$ \df{gives} $\alpha(k)=\mathrm{WER}^{pred}_k$, i.e. the regularization coefficient depends only on the sentence transcription quality. A value of $\beta=1$ gives $\alpha(k)=\beta$, i.e. the regularization coefficient remains  fixed  over all adaptation sentences (this is the case of Equation~\ref{eq:newp}). \df{Therefore, optimizing over $\beta$ allows us to control
the trade-off between the quality of the supervision and the size of the adaptation set.}

We refer to the DNN adaptation method based on Equation~\ref{eq:alfasent} as a ``soft'' adaptation \mn{(in which the coefficients vary sentence by sentence)}, in contrast with the ``hard'' DNN adaptation approach based on Equation~\ref{eq:newp} \mn{(in which the coefficients are fixed)}.

\section{ASR quality estimation}
\label{sec:quality}

The simplest approach to roughly estimate transcription quality (without reference transcripts) is to consider sentence confidence scores, which describe how the system is certain about the quality of its own hypotheses. Sentence confidence scores can be computed by averaging the confidence of the words in the best output string. 
Such information, however, often reflects a biased perspective influenced by individual ASR decoder features.

Indeed, confidence scores are usually close to the maximum value, 
thus shifting the predicted WER (computed as $1-confidence$) to scores that are close to zero. 

\begin{table*} [h]
\begin{center}
\begin{small}
\begin{tabular}{|p{1.7cm}|p{14cm}|}
\hline
\textbf{ASR} (9) &  From each CN bin: the log of the first word posterior (1), the log of the first word posterior from the previous/next bin (2), the mean/std/min/max of the log posteriors in the bin (4), if the first word of the previous/next bin is silence (2)   \\ \hline
\textbf{Sentence level} (10) & From each transcribed sentence: number of words (1), LM log probability (1), LM log probability of part of speech (POS) (1), log perplexity (1), LM log perplexity of POS (1),  percentage (\%)  of numbers (1), \% of tokens  which  do not contain only ``[a-z]'' (1), \% of content  words (1), \% of nouns (1), \% of verbs (1). \\ \hline
\textbf{Word level} (22) & From each transcribed word: Part-of-speech tag/score of the previous/current/next words (6), RNNLM   probabilities given by models trained on in-domain/out-of-domain  data (2), in-domain/out-of-domain 4-gram LM probability (2), number of   phoneme classes including fricatives, liquids, nasals, stops and vowels (5), number of homophones (1), number of lexical neighbors (heteronyms)
 (1) binary   features answering the three questions: ``is the current word a stop word?''/``is the current word before/after repetition?''/``is the current word before/after silence?'' (5). \\\hline
\end{tabular}
\end{small}
\end{center}
\caption{\label{feattab} 
Features (41 in total) for sentence-level WER prediction.}
\end{table*}

To obtain more objective and  reliable sentence-level WER predictions, in \cite{Negri:2014} we proposed ASR quality estimation 
as a supervised regression method that effectively  exploits a combination of ``glass-box'' and ``black-box'' features. Glass-box features, similar to confidence scores, 
capture information inherent to the inner workings of the ASR system that produced the transcriptions.
The black-box ones, instead, are extracted by looking only at the signal and the transcription. On one side, they try to capture the \textit{difficulty} of transcribing the signal while, on the other side, they try to capture the \textit{plausibility} of the output transcriptions. In both cases, the information used is independent from knowledge about the ASR system,  making 
ASR QE applicable to a wide range of scenarios in which the only  elements available for quality prediction are the signal and the transcription.

In this paper, we \df{trained} XRT-based models \cite{geurts2006extremely}  with a combination of 41 ASR (glass-box) and textual (black-box) features.  The ASR features are extracted from the confusion network (CN) \cite{mangu2000} 
derived from the word lattices generated by the ASR decoder (the one employed in this work is based on the KALDI toolkit \cite{Povey_ASRU2011}), while the textual features are the same of \cite{Negri:2014}. 
Table \ref{feattab} provides the complete list of the features used.
In our experiments, the regressors are respectively trained and tested on the CHiME-3  {\em dt05\_real} and  {\em et05\_real} transcriptions described in  Section~\ref{ssec:corpora}.
Their parameters, such as the number of bags, the number of trees per bag and the number of leaves per tree are  tuned to minimize the mean absolute error (MAE) between the true and predicted WER scores using k-fold cross-validation on the {\em dt05\_real} data.

\section{ASR system}
\label{sec:ASR}

\begin{figure}[!h]
\centering
\includegraphics[width=16.0cm, angle=0]{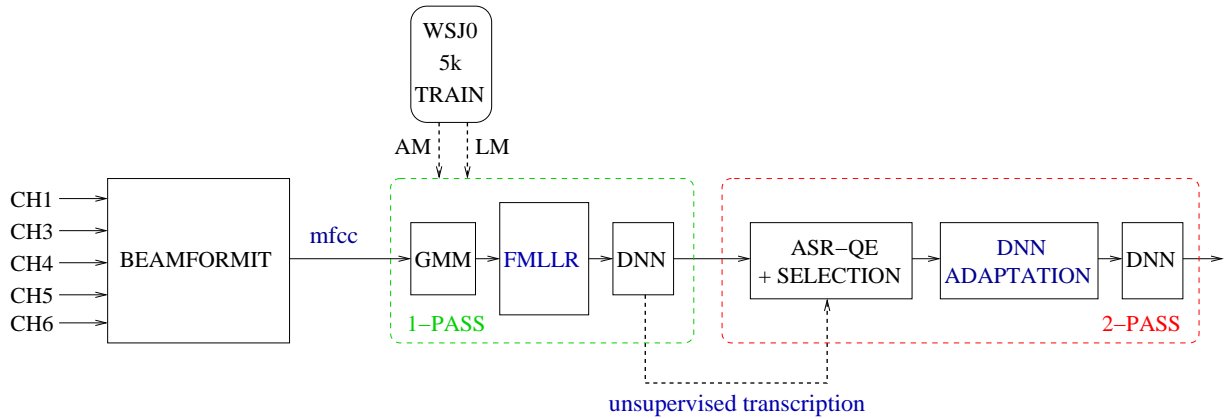}
\caption{ASR architecture based on the KALDI CHiME-3 v2 package plus ASR QE hypotheses selection and DNN adaptation.}
\label{fig:baseline_fmllr}
\end{figure}

\begin{figure}[!h]
\centering
\includegraphics[width=16.0cm, angle=0]{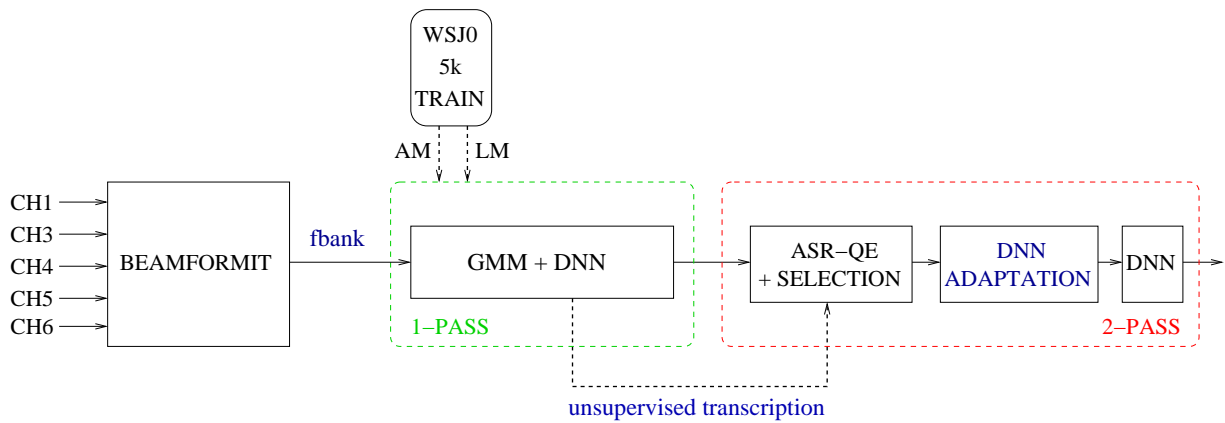}
\caption{ASR architecture based on the KALDI CHiME-3 package based on standard filter-bank features. 
}
\label{fig:baseline_fbank}
\end{figure}

The architecture of our ASR system  is depicted in Figures~\ref{fig:baseline_fmllr} and \ref{fig:baseline_fbank}. In the former one, it uses  fMLLR normalized features; in the latter one, it uses  filter-bank features.
The system is mainly based on the KALDI CHiME-3 v2 package \df{(derived from the ASR system described in \cite{hori2015})} with the addition of a second decoding pass that performs unsupervised DNN adaptation 
as described in Section~\ref{sec:adaptation}. 

In our submission to CHiME-3 \cite{falavi2015}, we reached the best performance on the evaluation set,  {\em et05\_real}, with a simple delay-and-sum (DS) beamforming consisting in uniform weighting of the rephased signals of the 5 frontal microphones. A similar approach, although based on the well known BeamformIt toolkit \cite{anguera2007}, is also included in the recent software package implementing the CHiME-3 baseline. Hence, in order to comply with the baseline, for this work we used  BeamformIt to implement signal enhancement.

After beamforming, both  filter-bank and fMLLR features are computed and processed by a corresponding hybrid DNN-HMM system that produces the supervision for adapting the DNN  
in the final decoding pass.

\subsection{Filter-bank features}

The employed filter-bank consists of 40 log Mel scaled filters. Feature vectors are computed every 10{\em ms} by using a Hamming window of 25{\em ms} length and are    mean/variance   normalized  on a speaker-by-speaker basis.
The baseline DNN is trained using the Karel's setup \cite{karel2011} included in the KALDI toolkit. To this aim the 
8,738 training utterances were aligned to their transcriptions by means of the baseline GMM-HMM models.\footnote{The initial GMM system makes use of the KALDI recipe associated to the earlier CHiME challenges \cite{CHiME1,CHiME2}.} An $11$-frame context window (5 frames on  each side) is used as input to form a 440  dimensional feature vector.  The DNN has 7 hidden layers, each with 2,048 neurons.  The DNN  is  trained  in  several stages  including  Restricted  Boltzmann Machines  (RBM) pre-training,  mini-batch  stochastic gradient  descent training, and sequence-discriminative training using state-level Minimum Bayes Risk (sMBR). Initially,  the learning rate is set to 0.008 and it is halved every time the relative difference in frame accuracy between two epochs on a cross-validation set falls below 0.5\%.  A frame accuracy improvement on the cross-validation set lower than 0.1\% stops the optimization. 
All experiments involving adaptation of the baseline DNN,  aimed at minimizing the objective function defined in Equation~\ref{eq:div1},  are performed according to the above recipe.

\subsection{fMLLR features}

For training, 13 mel-frequency cepstral coefficients  (MFCCs) are computed every 
10{\em ms} by using a Hamming window of 25{\em ms} length. These  features  are    mean/variance   normalized  on
a speaker-by-speaker basis, spliced by +/-  3 frames next to the central frame and  projected down to  40 dimensions using  linear discriminant analysis  (LDA) \df{and maximum likelihood linear transformation (MLLT) \cite{Povey_ASRU2011}. Then, a single speaker-dependent fMLLR transform is  estimated  and applied for speaker adaptive training of triphone HMM-GMMs.
The DNN-HMMs hybrid systems are built on top of LDA+MLLT+fMLLR features and SAT triphone HMM-GMMs.}

\df{During decoding, first LDA+MLTT+fMLLR are derived using auxiliary HMM-GMMs. To this end, a preliminary decoding pass with speaker-independent (SI) HMM-GMM is conducted to produce a word lattice for each input utterance.} Then, a single  fMLLR transform for  each speaker  is estimated from sufficient statistics collected from SI word lattices \df{in order to maximize the likelihood of the acoustic observations given the  SAT triphone HMM-GMMs.} These  transforms are 
used  with  SAT triphone HMM-GMMs to  produce new word  lattices. A second set of fMLLR transforms is estimated from new word lattices and combined with the  first set of transforms. Finally, the resulting transforms are   applied to normalize the features processed by the DNN-HMM hybrid system in the first decoding pass of Figure~\ref{fig:baseline_fmllr}.
The training of the corresponding baseline DNN, as well as DNN adaptation by KLD regularization, 
use 
the  recipe adopted for filter-bank features.

\subsection{Language models}

The  LM employed in the experiments is the 3-gram LM provided with the CHiME-3 v2 package release, which uses the Kneser-Ney smoothing method for estimating back-off probabilities. It was trained with around 37 million words. After pruning  low frequency words, the vocabulary size is approximately 5,000 words and the perplexity value (measured over {\em dt05\_real} reference transcriptions) is 119.2.

Finally, although not depicted in the figure, we also run a final rescoring of the n-best lists generated in the second decoding  pass with the 5-gram LM and the RNNLM included in the  CHiME-3 v2 package.

\section{Experimental setup}
\label{sec:setup}

\subsection{Speech corpora}
\label{ssec:corpora}

For our experiments, we use the multiple-microphone evaluation data collected for the CHiME-3 challenge, which is publicly available.\footnote{\url{http://spandh.dcs.shef.ac.uk/chime_challenge/download.html}.} Complete details about this data set, the overall challenge and its outcomes can be found in the related overview paper \cite{CHiME3}, which also reports the performance of the 26 participating systems.

Six different microphones, placed on a tablet PC, were used to record sentences of the Wall Street Journal (WSJ) corpus, uttered by different speakers in four different environments (bus, cafe, pedestrian area and street junction). The training corpus consists of 1,600 ``real'' noisy sentences  uttered by 4 speakers, and of 7,138 ``simulated'' noisy sentences uttered by 83 speakers forming the WSJ SI-84 training set. Simulated noisy sentences are generated by means of convolution of clean signals with impulse responses  of the above mentioned environments and summing the corresponding pre-recorded background noises.

Two evaluation corpora were collected in this scenario: the {\em dt05\_real} development set, formed by 1,640 sentences uttered by 4 different speakers, and the {\em et05\_real} test set, formed by 1,320 utterances  acquired from 4 other speakers. In addition, two parallel sets of ``simulated'' noisy utterances (namely {\em dt05\_simu} and {\em et05\_simu}) were generated as previously described. There is no speaker overlap between training, development and test sets. The number of utterances  in the evaluation corpora is equally distributed among speakers and types of noise, that is, every speaker uttered the same number of sentences in each of the four environments. In both training and evaluation data sets, utterance segmentation  was manually checked and the corresponding speaker identity was annotated. Therefore, no automatic speaker diarization module  was employed in the experiments. Table \ref{chimestat} shows some statistics of the CHiME-3 training, 
\textit{real} development and  test sets 
(the corresponding simulated development and test sets exhibit the same statistics of the \textit{real} data). 

\begin{table} [h]  
\begin{center}
\begin{tabular}{|l|l|l|l|l|}
\hline
& tr05\_simu & tr05\_real &dt05\_real & et05\_real  \\ \hline
duration & 15h9m & 2h54m & 2h16m & 1h50m \\
\# sentences & 7,138 & 1,600 & 1,640 & 1,320 \\
\# words & 136.5k & 28.3k & 27.1k & 21.4k  \\
dict. size & 8.9k & 5.6k & 1.6k & 1.3k \\
\# speakers & 83 & 4 & 4 & 4 \\
\# noises & 4 & 4 & 4 &4 \\\hline
\end{tabular}
\end{center}
\caption{\label{chimestat}  Statistics of CHiME-3 training, real development and real test audio data.}
\end{table}

\subsection{Experiment definition}
\label{ssec:expe}
All the experiments  were conducted on the ``real'' subsets of the CHiME-3 evaluation data: {\em dt05\_real} and {\em et05\_real}.
For brevity, henceforth we will respectively refer to them as \textit{DT05} and \textit{ET05}. We report performance for   ASR systems employing both fMLLR normalized features (Figure~\ref{fig:baseline_fmllr}) and filter-bank features (Figure~\ref{fig:baseline_fbank}).
ASR parameters (LM weight, $\alpha$ and $\beta$  coefficients in Equations~\ref{eq:newp} and \ref{eq:alfasent} respectively) are tuned on the development set \textit{DT05}.

The soft adaptation approach described in Section~\ref{sec:adaptation} was applied in both ``oracle'' and ``predicted'' conditions. Oracle WER scores (\textit{oWER} henceforth) are computed from 
reference transcriptions, while predicted WER scores (\textit{pWER}) are estimated by the ASR QE system described in Section~\ref{sec:quality}. Both  values are used as WER estimates in Equation~\ref{eq:alfasent} to compute the target probability distribution. The performance achieved by oracle sentence WER represents the upper bound of the soft adaptation approach.

The QE model used for WER prediction is trained and optimized on the development set. The XRT parameters are tuned in 8-fold cross validation, minimizing  the mean absolute error (MAE) between  the predicted   and the
true WER. The partitioning is done to avoid speaker or sentence overlaps between  training and test 
folds. 

Table \ref{table:expeall} gives the complete list of DNN adaptation experiments we performed. Each experiment is identified by: {\em i)}  a 
combination of adaptation/evaluation sets, {\em ii)} the supervision used (manual or automatic), and {\em iii)} the features employed (filter-bank or fMLLR normalized). For instance, the experiment named {\em DT05+man+fMLLR+ET05} in the first row of the table indicates that the baseline DNN is adapted using {\em DT05} as adaptation set, the manual supervision, fMLLR   features and the evaluation set is {\em ET05}.

\begin{table}[!h]
\begin{center}
\begin{tabular}{|l|c|c|c|c|} 
\cline{2-5}
\multicolumn{1}{c|} { } & {adaptation} & {type  of} & {features} & {evaluation}  \\
\multicolumn{1}{c|} {\df{cross conditions}  } & {set} & {supervision} & {type} &  {set}  \\\hline\hline
DT05+man+fMLLR+ET05 & DT05 & manual  & fMLLR& ET05 \\\hline
DT05+man+fbank+ET05 & DT05 & manual  & filter-bank & ET05 \\\hline
DT05+auto+fMLLR+ET05 & DT05 & automatic  & fMLLR& ET05 \\\hline
DT05+auto+fbank+ET05 & DT05 & automatic  & filter-bank & ET05 \\\hline\hline
\multicolumn{1}{c|}{\df{homogeneous conditions} } & { } & { } & { } & { }\\\hline\hline
DT05+auto+fMLLR+DT05 & DT05 & automatic  & fMLLR & DT05 \\\hline
DT05+auto+fbank+DT05 & DT05 & automatic  & filter-bank & DT05 \\\hline
ET05+auto+fMLLR+ET05 & ET05 & automatic & fMLLR & ET05 \\\hline
ET05+auto+fbank+ET05 & ET05 & automatic & filter-bank & ET05 \\\hline\hline
\end{tabular}
\caption{List of DNN adaptation experiments.}
\label{table:expeall}
\end{center}
\end{table}

\mn{The table is divided in two parts, respectively describing experiments carried out in \textit{cross} and \textit{homogeneous} conditions. In cross conditions (first four rows), the adaptation and evaluation sets are distinct. Here, our goal is to compare performance by varying the type of supervision (manual or automatic) of the adaptation data  and, in case the automatic supervision is used, by varying the size of the adaptation set according to its quality. In homogeneous conditions (last four rows), the adaptation set coincides with the evaluation set. In this case, our goal is to compare performance achieved by selecting adaptation sets with different levels of quality. }

Note that DNN adaptation with manual supervision (first two rows of Table \ref{table:expeall}) is only meaningful in cross conditions, since we assume it is not available for the evaluation set {\em ET05}.
The automatic supervisions of the adaptation sets (i.e {\em DT05} or {\em ET05}, depending on the experiment type)
are produced by the first decoding passes 
of the ASR systems depicted in
 Figures~\ref{fig:baseline_fmllr} 
and~\ref{fig:baseline_fbank}.
KLD regularization with manual supervision is applied 
according to Equation~\ref{eq:newp}. Instead, with automatic supervision, both hard (based on Equation~\ref{eq:newp}) and soft (based on Equation~\ref{eq:alfasent}) DNN adaptation approaches are applied.

\df{Furthermore, it's worth observing that the cross-condition situation fits  an ``offline'' application scenario, where a DNN can be adapted using data and corresponding automatic transcriptions collected on the field during the ASR system working. In a successive phase, the adapted DNN can be  loaded  into the ASR system itself.}

Finally, as mentioned in 
Section~\ref{sec:intro},
and as it will be shown below, top performance is achieved 
by properly selecting subsets of the 
adaptation data. 
Therefore, similarly to the other tuning parameters, the optimal  selection thresholds used to estimate the final performance are computed  on the  development set {\em DT05}.

\section{Results}
\label{sec:results}

\mn{In this section we present the experimental results obtained in the different settings 
outlined in Table~\ref{table:expeall}, playing with: \textit{i)} the different conditions, \textit{ii)} the type of supervision, \textit{iii)} the size of the} \df{adaptation data} \mn{and \textit{iv)} the way the adaptation data is selected.}

\mn{In the analysis and in the subsequent discussion, our WER scores are not compared against those achieved by the
best ASR system  participating  in the CHiME-3 challenge \cite{yoshioka2015}, which uses a far more complex architecture for  signal pre-processing and cross system combination,   as well as an augmented set of training data. 
%
Indeed, implementing such a state-of-the-art system was  out of the scope of this work, whose objective is to show the effectiveness of 
QE-based DNN adaptation
to improve the performance of a standard,  less complex but still strong ASR system.
For this reason, our term of comparison is represented by the reference CHiME-3 baseline, which results in 15.4\% WER on {\em ET05} and 8.2\% WER on {\em DT05}.
Despite the generality of the proposed approach, integrating our method
in 
a state-of-the-art \df{ASR system like the one described in  \cite{yoshioka2015}} and quantify the performance gains yielded by QE-based DNN adaptation is left as a possible direction for future work.}

\subsection{DNN adaptation in cross conditions}

\mn{In  this  section  we  analyse  the  performance  achieved in cross conditions, both   with manual and automatic supervision. First, we use use all the sentences in  {\em DT05} for adapting the DNN. Then,  we show the performance achieved with the automatic supervision provided by  adaptation data derived  from {\em DT05} after removing the utterances with the highest WER.}

\subsubsection{Using all the adaptation utterances}
\label{subsubsec:all1}

\begin{figure}[!h]
\centering 
\begin{subfigure}[b]{0.5\textwidth}
\includegraphics[width=5.8cm, angle=-90]{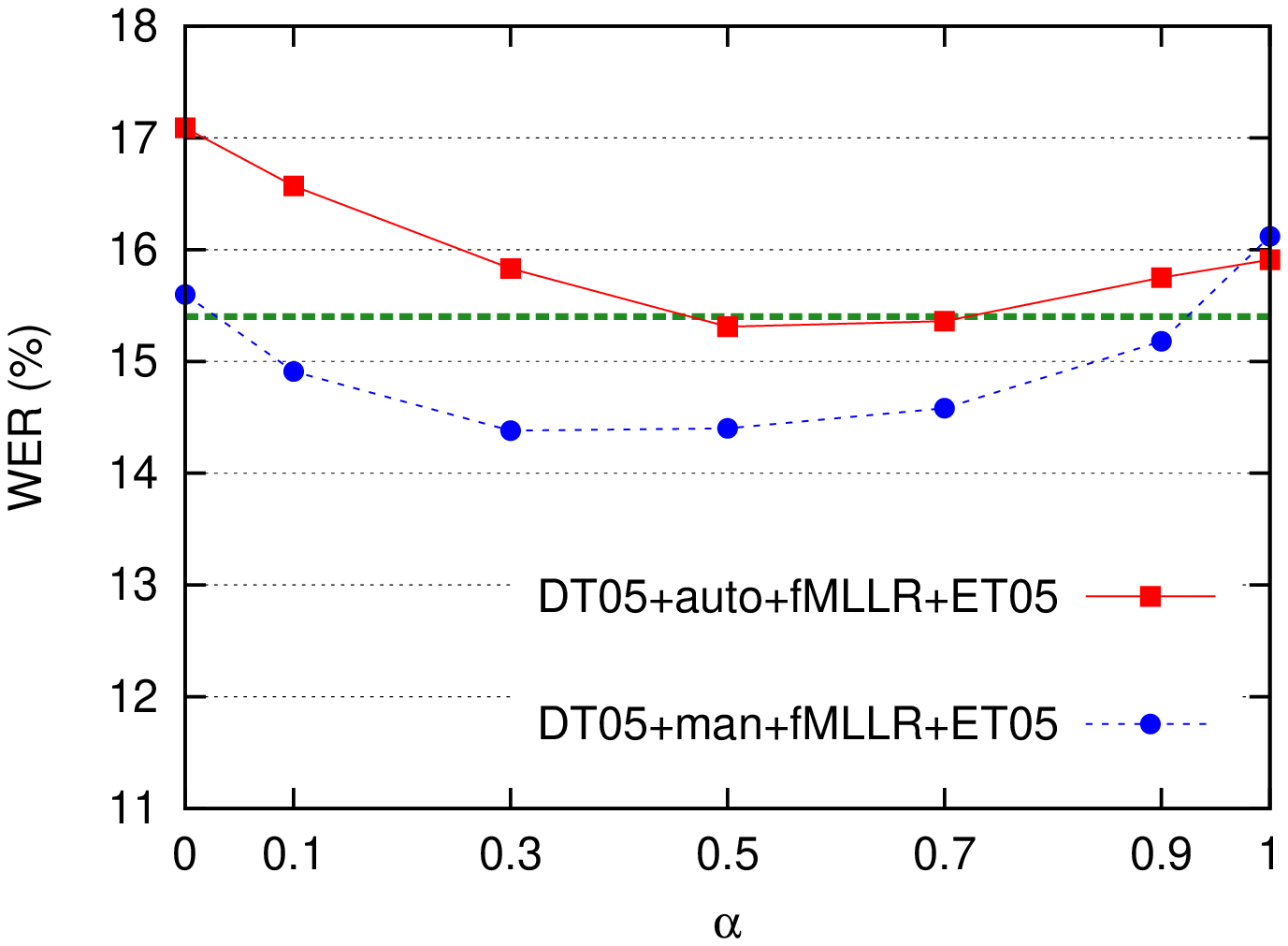}
\caption{}
\end{subfigure}
\begin{subfigure}[b]{0.48\textwidth}
\includegraphics[width=5.8cm, angle=-90]{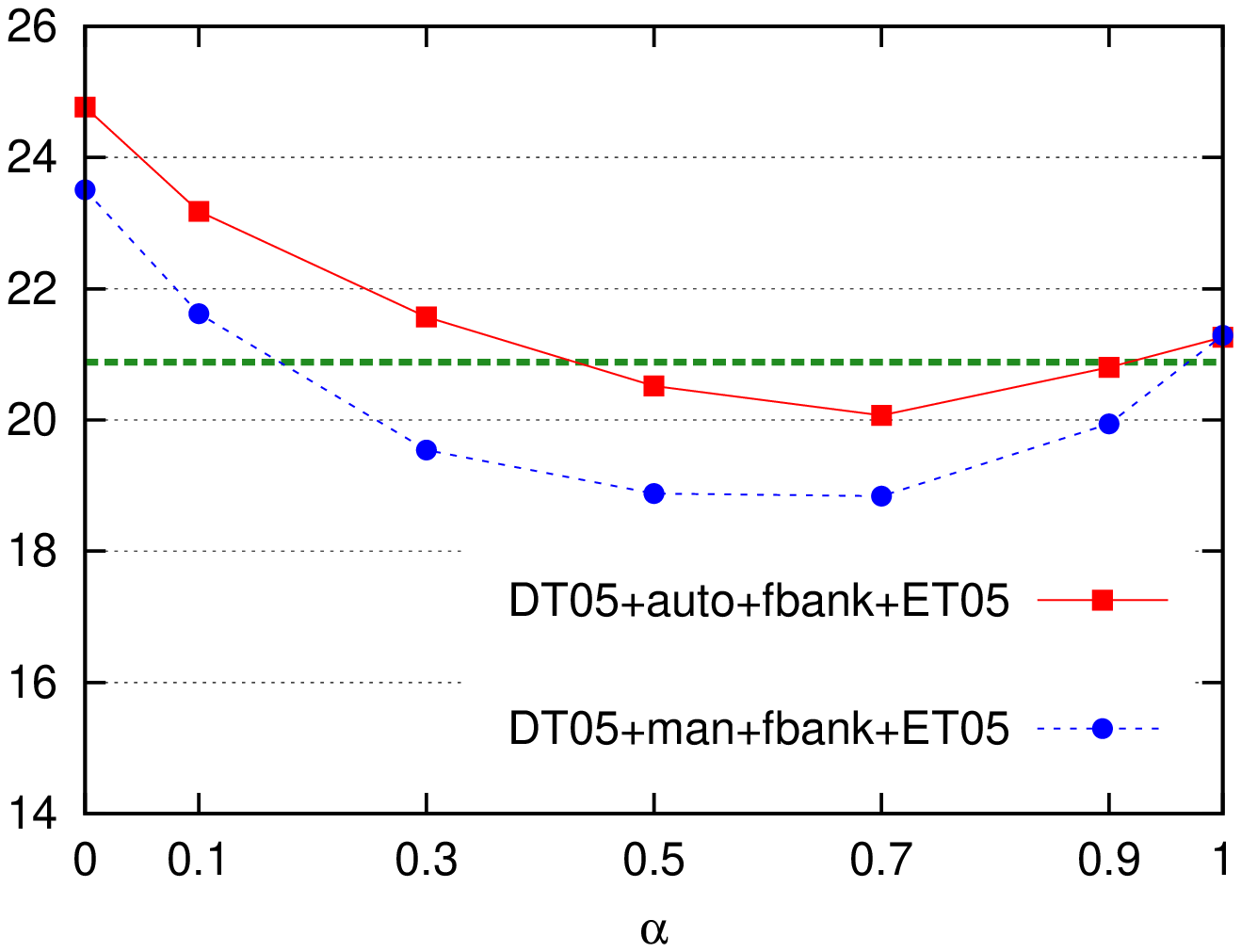}
\caption{}
\end{subfigure}
\caption{WER achieved on evaluation set {\em ET05} as a function of the regularization coefficient $\alpha$, using {\em DT05} as adaptation set.}
\label{fig:dt05_real_sup_et05_real_fmllr}
\end{figure}

\mn{Figure~\ref{fig:dt05_real_sup_et05_real_fmllr}a shows the WERs (as functions of the regularization coefficient $\alpha$ in Equation~\ref{eq:newp}) achieved on the evaluation  set {\em ET05} by using fMLLR features both with  manual and automatic supervision. In a similar way, the performance reached with filter-bank features is given in Figure~\ref{fig:dt05_real_sup_et05_real_fmllr}b. The horizontal line in both the figures corresponds to the baseline performance.}

\mn{As can be seen, the use of  manual supervision, or equivalently  the  {\em supervised adaptation}, allows to improve baseline performance  with both types of features.}
In both cases, there is an intermediate  optimal value of $\alpha$ in the interval $[0,1]$, indicating that we should not totally trust neither  the original model nor the adaptation data.\footnote{As explained in Section~\ref{sec:adaptation}, a value of $\alpha=0$ corresponds to completely ignoring the contribution of the original DNN output distribution in the construction of the cross-entropy function (i.e. we completely trust the adaptation data), while a value $\alpha=1$ forces the DNN parameters to follow those of the original distribution (i.e. we completely trust the original model).} With the best value, we gain about 1\% WER point, indicating the efficacy of the interpolation procedure expressed by Equation~\ref{eq:newp}.

Note the substantial performance reduction 
in Figure~\ref{fig:dt05_real_sup_et05_real_fmllr}b at $\alpha=0$ (both for supervised and unsupervised adaptation), suggesting that a data overfitting effect has probably occurred. The same behavior is not observed with  supervised adaptation using fMLLR features 
(Figure~\ref{fig:dt05_real_sup_et05_real_fmllr}a), where at $\alpha=0$ no significant performance degradation is observed. 
This result can be explained by considering that fMLLR transformations 
\mn{already reduce}
the acoustic mismatch between adaptation ({\em DT05}) and evaluation ({\em ET05}) sets. 
This is confirmed by the fact
that, in Figure~\ref{fig:dt05_real_sup_et05_real_fmllr}b, the  curve labelled {\em DT05+man+fbank+ET05} is shifted towards the right part of the graph more than the corresponding curve {\em DT05+man+fMLLR+ET05} in Figure~\ref{fig:dt05_real_sup_et05_real_fmllr}a, 
meaning that the adaptation procedure trusts  the fMLLR normalized features more than the filter-bank ones.
Referring to the same Figure~\ref{fig:dt05_real_sup_et05_real_fmllr}a, data overfitting (at $\alpha=0$) instead occurs with unsupervised adaptation, as if the errors in the supervision acted similarly to an acoustic mismatch between adaptation and evaluation sets. 
Based on these outcomes,
we decided to investigate the effects of reducing the errors in the automatic transcription.

\subsubsection{Selecting adaptation utterances}

\begin{figure}[!h]
\centering
\begin{subfigure}[b]{0.5\textwidth}
{\includegraphics[width=5.8cm,angle=-90]{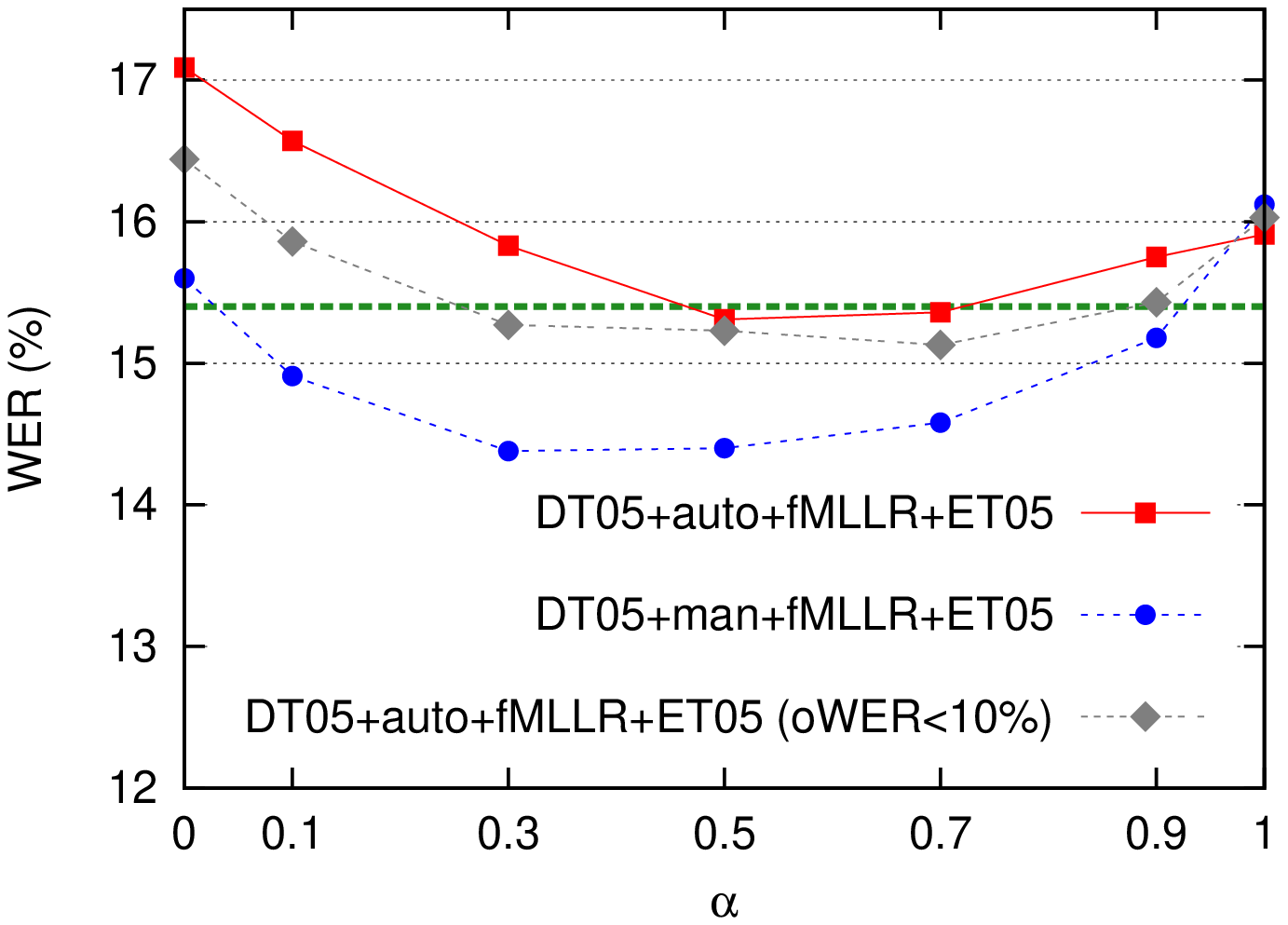}}
\caption{}
\end{subfigure}
\begin{subfigure}[b]{0.48\textwidth}
{\includegraphics[width=5.8cm,angle=-90]{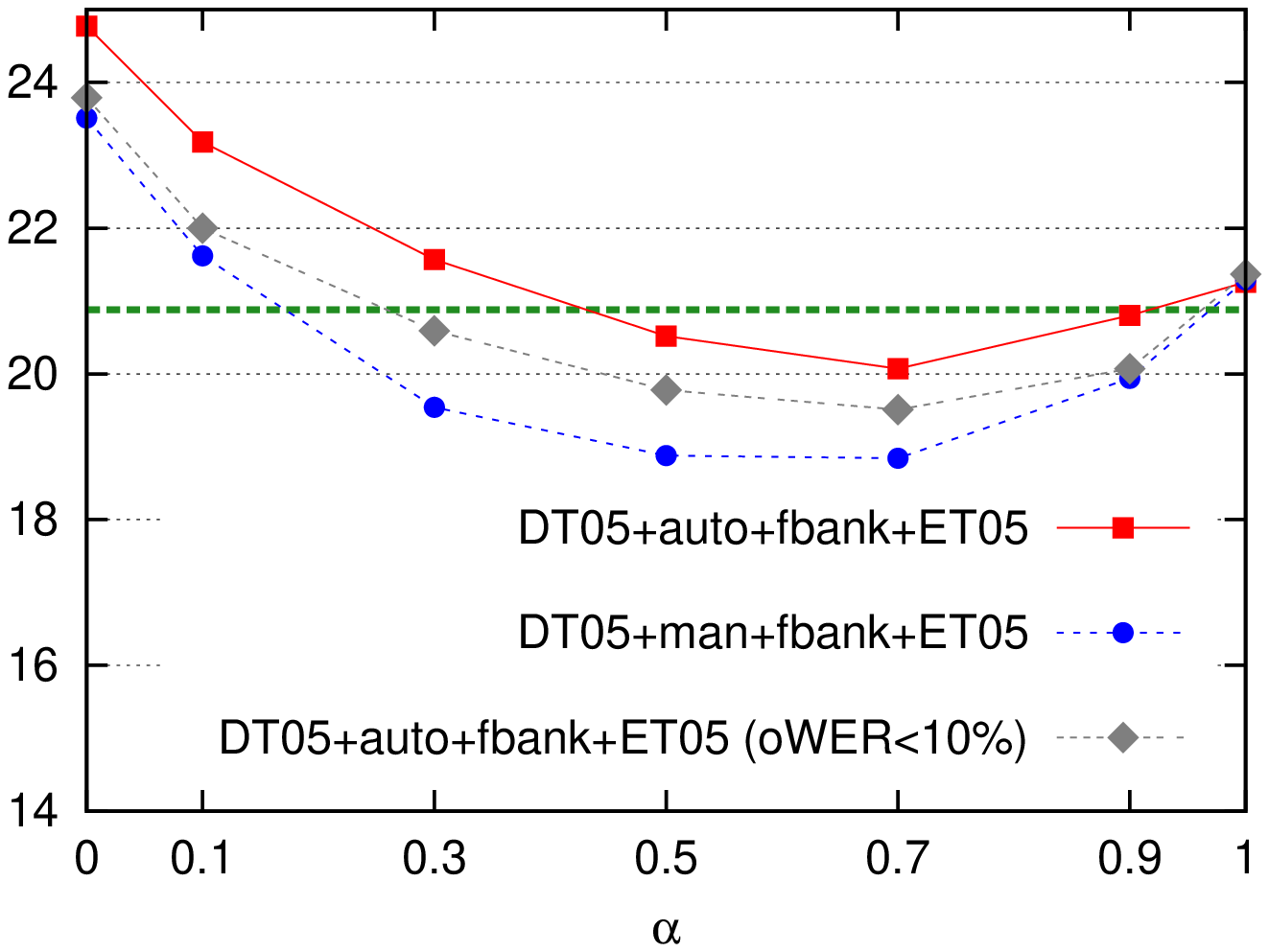}}
\caption{}
\end{subfigure}
\caption{WER achieved on evaluation set {\em ET05} as a function of the regularization coefficient $\alpha$, using as adaptation set the subset of {\em DT05} with $oWER\le$10\%.}
\label{fig:dt05_real_unsupwer0_et05_real}
\end{figure}

To check the possible impact of automatic transcription errors in the supervision,
we extracted from the adaptation set {\em DT05} the utterances whose  true WER, computed from the reference transcriptions, is lower than 10\%. Then, we adapted the baseline DNN with  the hard approach and by varying the value of the regularization coefficient $\alpha$. 
The 
results are shown in  Figures~\ref{fig:dt05_real_unsupwer0_et05_real}a and~\ref{fig:dt05_real_unsupwer0_et05_real}b. For comparison purposes, the two figures also include the same curves of Figures~\ref{fig:dt05_real_sup_et05_real_fmllr}a and~\ref{fig:dt05_real_sup_et05_real_fmllr}b
related to the use of the whole adaptation set.
%
As can be seen, the selection of  adaptation utterances with WER$<10\%$ produces curves that approach those obtained using 
 manual supervision,
showing 
the benefits of reducing the transcription errors in the supervision.

\subsection{DNN adaptation in homogeneous  conditions}

\mn{In this  section we report and discuss the performance achieved in homogeneous conditions with automatic supervision. First, we use the whole available set of adaptation utterances. 
Then, by applying  ASR QE for WER prediction, we experiment with different subsets of the data selected according to their estimated quality.}

\subsubsection{Using  \textit{all} the adaptation utterances}
\label{subsubsec:all2}

The experiments  were conducted 
 by performing
 two   decoding passes,\footnote{These experiments were motivated by the significant performance improvement obtained in \cite{falavi2015} using ``full'' retraining of DNN in a two-pass ASR architecture.} as explained in Section~\ref{sec:ASR}. 
The transcriptions resulting from the first pass, based on  the baseline DNNs, provide the supervision for the 
following adaptation steps. Then, the adapted DNNs are exploited in the second decoding pass to produce the final transcriptions. 
%
Performance results achieved on both  development and evaluation sets, with hard and soft adaptation, are given in Table~\ref{table:expeunsup} (in parentheses, we show the absolute WER reduction with respect to baseline performance).
\begin{table}[!h]
\begin{center}
\begin{tabular}{|l|c|c|c|} 
\hline
experiment code & HARD  & SOFT ({\em oWER}) & SOFT ({\em pWER})\\
& \df{adaptation} & \df{adaptation} & \df{adaptation} \\\hline 
\hline
DT05+auto+fMLLR+DT05 & 8.0(0.2) & 7.9(0.3) & 8.0(0.2)   \\\hline
DT05+auto+fbank+DT05 & 9.5(1.6) & 9.2(1.9) & 9.3(1.8)   \\\hline
ET05+auto+fMLLR+ET05 & 14.5(0.9) & 14.3(1.1)&14.4(1.0) \\\hline
ET05+auto+fbank+ET05 & 17.7(3.2) & 17.1(3.8)&17.6(3.3) \\\hline
\end{tabular}
\caption{\%WER achieved by unsupervised DNN adaptation in homogeneous conditions.}
\label{table:expeunsup}
\end{center}
\end{table}
In the case of soft adaptation, both oracle and automatically-predicted sentence WERs were tested. Similarly to the experiments in Figure~\ref{fig:dt05_real_sup_et05_real_fmllr}, we measured performance as a function of the coefficient $\alpha$. \df{We also carried out the same set of experiments evaluating performance also as function of  coefficient $\beta$ defined in Equation~\ref{eq:alfasent}.}
However, for reasons \df{of compactness}, we do not provide the whole set of results and Table~\ref{table:expeunsup} only refers to the top WER values achieved.

\begin{figure}[!h]
\centering
\begin{subfigure}[b]{0.48\textwidth}
{\includegraphics[width=7.5cm,angle=-90]{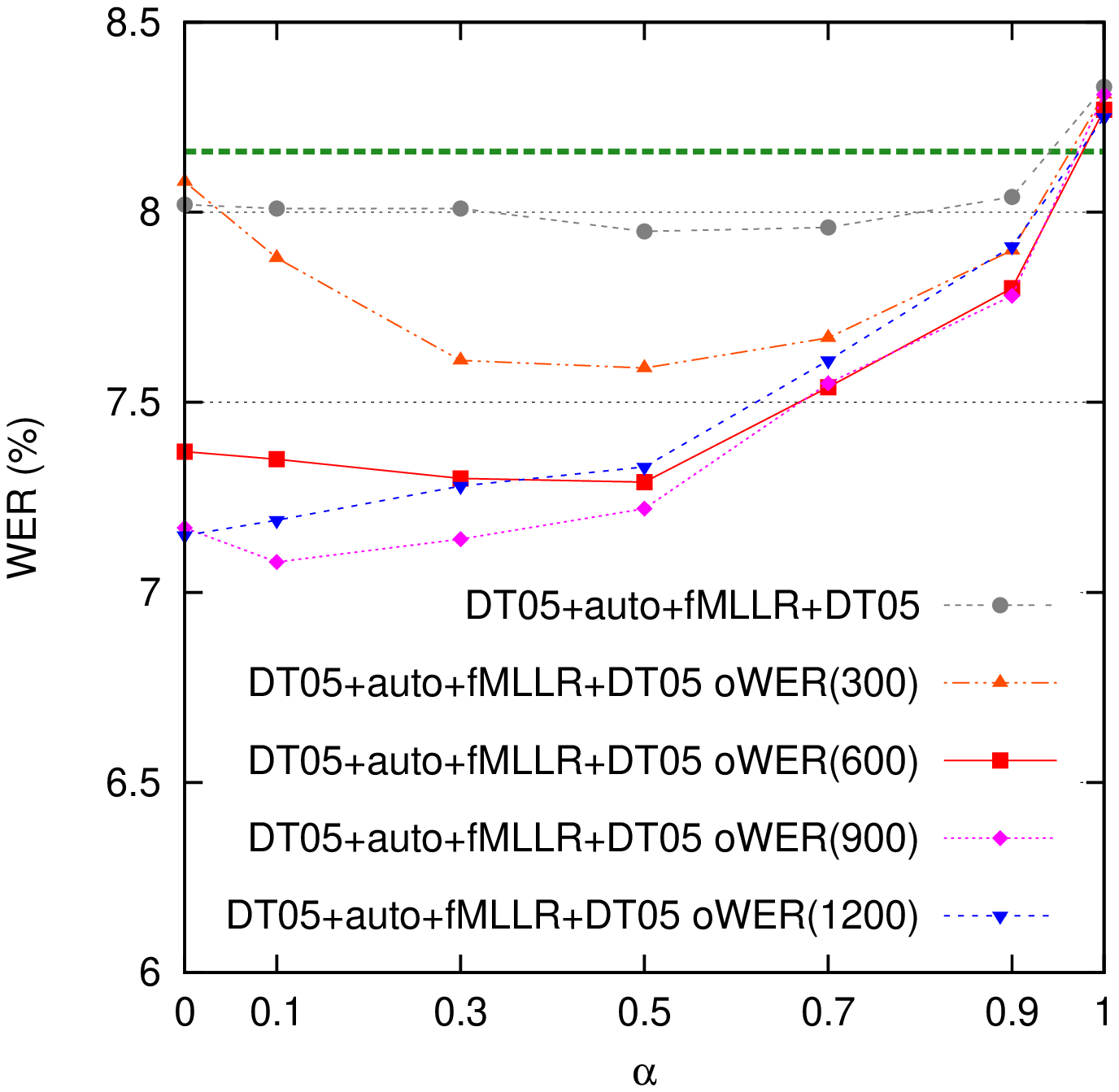}}
\caption{}
\end{subfigure}
\begin{subfigure}[b]{0.48\textwidth}
{\includegraphics[width=7.5cm,angle=-90]{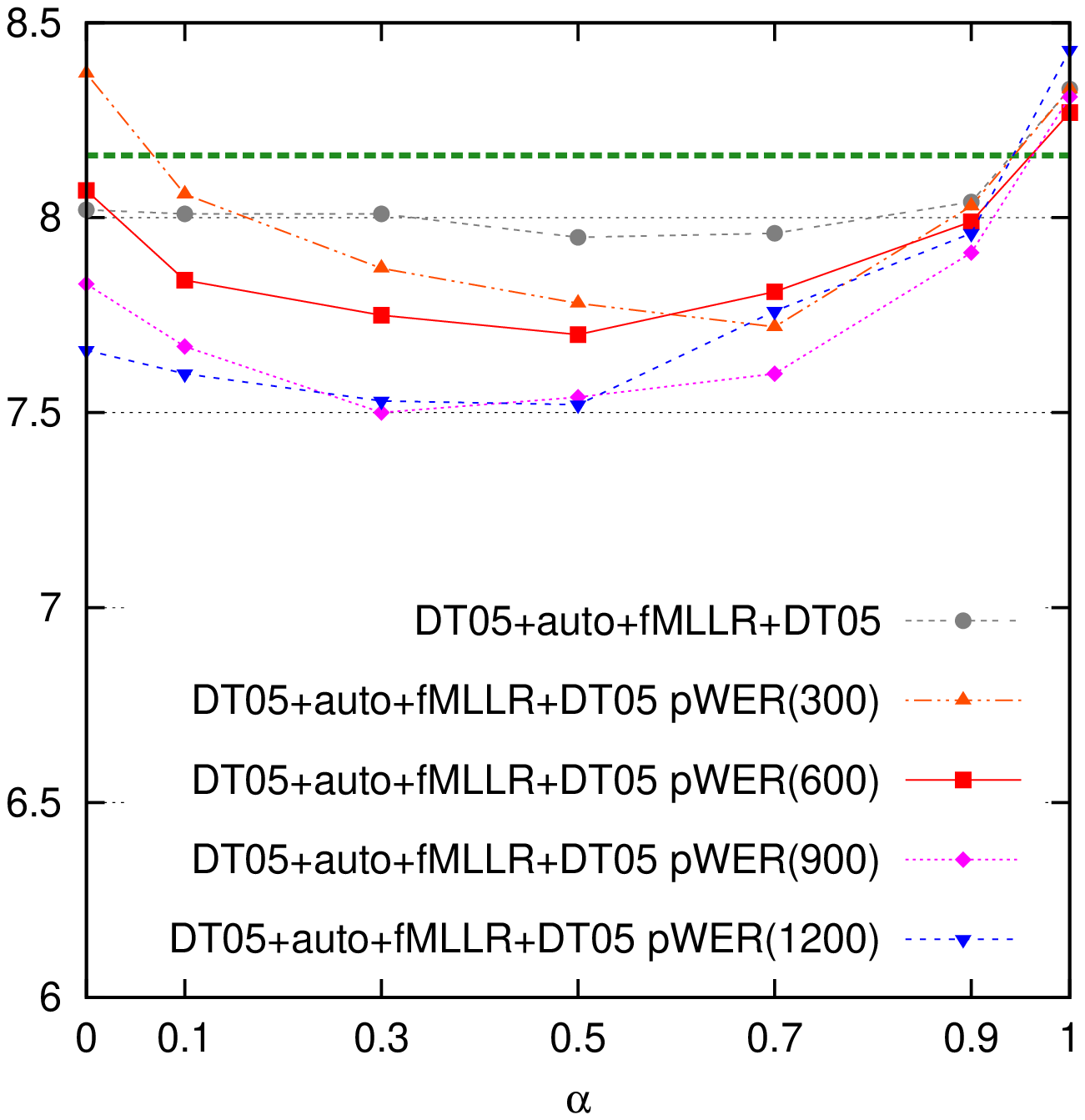}}
\caption{}
\end{subfigure}
\caption{WER achieved with oracle ({\em oWER}) and ASR QE ({\em pWER}) selection of adaptation utterances, on the development set {\em DT05}, as functions of the regularization coefficient $\alpha$.}
\label{fig:dt05_real_oracle-predicted_vs_adasize}
\end{figure}

First of all, differently from  the performance  
shown in Figures~\ref{fig:dt05_real_sup_et05_real_fmllr}a and ~\ref{fig:dt05_real_sup_et05_real_fmllr}b,  experiments in homogeneous conditions do not exhibit clear minimum values of the corresponding WER. Basically, no significant WER variations are observed for both $\alpha$ and $\beta$ coefficients ranging in the interval $[0.0-0.7]$. The best performance is achieved for $\alpha=\beta=0.7$, while for $(\alpha,\beta)>0.7$ the WERs increase.\footnote{To see examples of this trend of performance the reader can refer to the  Figures~\ref{fig:dt05_real_oracle-predicted_vs_adasize}a,~\ref{fig:dt05_real_oracle-predicted_vs_adasize}b,~\ref{fig:et05_real_oracle-predicted_vs_adasize}a and ~\ref{fig:et05_real_oracle-predicted_vs_adasize}b,
which report the WER scores achieved with hard adaptation and fMLLR features in homogeneous conditions (specifically, refer to the curves obtained without automatic sentence selection, respectively {\em DT05+auto+fMLLR+DT05} and {\em ET05+auto+fMLLR+ET05}).}

In Table~\ref{table:expeunsup} it is worth noting the significant WER reductions, compared to  baseline performance, yielded by filter-bank features on both {\em DT05} and {\em ET05}. Although similar performance gains are not observed with fMLLR features, especially on {\em DT05} (as just pointed out above, probably due to their capability of reducing the acoustic mismatch between training and testing conditions), 
these results confirm
the effectiveness of the two-pass decoding method. Note also that no substantial advantages 
are brought by 
the soft adaptation approach compared to the hard one. Despite this fact, it is worth observing  
the  very close performance  between  oracle and predicted WER estimates, which demonstrates the efficacy of the proposed ASR QE approach.

In summary, what one can learn from 
the experiments discussed so far 
is that: {\em i}) DNN adaptation in homogeneous conditions with two passes of decoding, using the whole set of adaptation utterances, yields performance improvements, {\em ii}) \df{automatic utterance selection} based on oracle WER values is effective in cross conditions suggesting, together with the outcome above,   to repeat the corresponding experiment in homogeneous conditions and \df{using ASR QE}, and {\em iii}) no significant performance gain can be achieved with the 
soft adaptation method based on Equation~\ref{eq:alfasent}. 
\df{The latter result suggests to investigate measures for expressing sentence quality that are alternative  to sentence WER used in Equation~\ref{eq:alfasent}. In addition, weighing of output probabilities in Equation~\ref{eq:newp}, could be applied at a granularity level higher than that of the sentence, e.g. at the level of word or even of single frame. Though interesting, a formal verification of these hypotheses is out of the scope of this paper and is left for future work.}

\subsubsection{Selecting adaptation utterances}

For 
conciseness, 
in the next set of experiments we report performance results only for fMLLR normalized features, since they are the most effective ones. However, the same and even more evident trends,  were also observed using filter-bank features.
Figures~\ref{fig:dt05_real_oracle-predicted_vs_adasize}a and~\ref{fig:dt05_real_oracle-predicted_vs_adasize}b 
report the performance achieved on {\em DT05} using subsets of adaptation utterances of different size.
The  utterances of the development set {\em DT05} were sorted according to the WER resulting from the first decoding pass. For sorting, we used 
 both oracle WER values and WER predictions obtained with the ASR QE approach described in Section~\ref{sec:quality}.
We extracted from {\em DT05} four adaptation sets, respectively containing the ``best'' $300$, $600$, $900$ and $1,200$ utterances. The various subsets, together with their automatic transcriptions, were used to adapt the baseline DNN by means of the hard approach. The reason for putting thresholds to the size of the adaptation set to compute
our results
lies in the fact that we want to do a fair comparison between the two  selection methods adopted ({\em oWER} and {\em pWER}).
In fact, sentence selection according to a preassigned WER threshold produces unbalanced adaptation sets of different sizes in correspondence to the application of each of the two  methods.
\begin{figure}[!h]
\centering
\begin{subfigure}[b]{0.48\textwidth}
{\includegraphics[width=7.5cm,angle=-90]{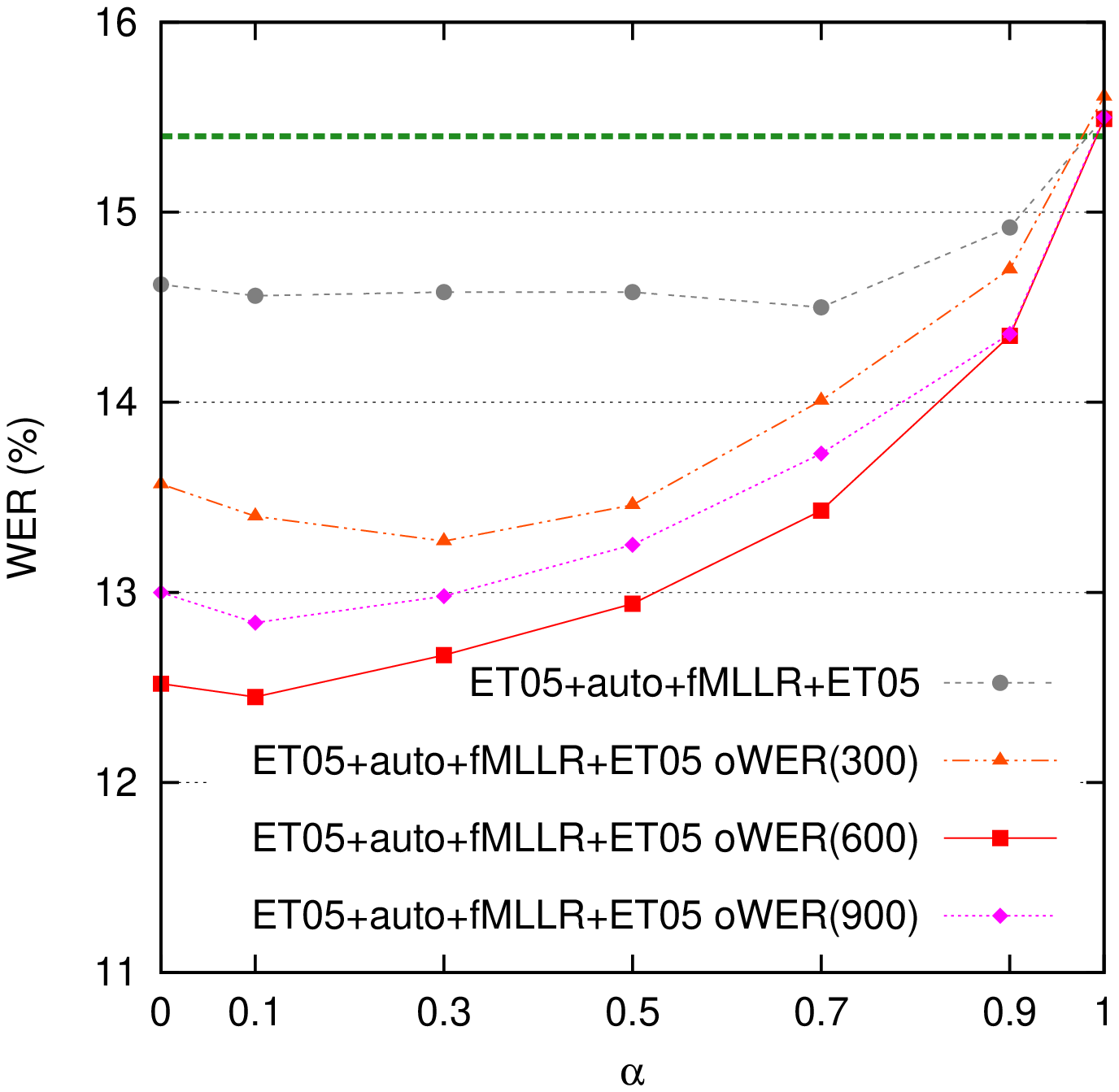}}
\caption{}
\end{subfigure}
\begin{subfigure}[b]{0.48\textwidth}
{\includegraphics[width=7.5cm,angle=-90]{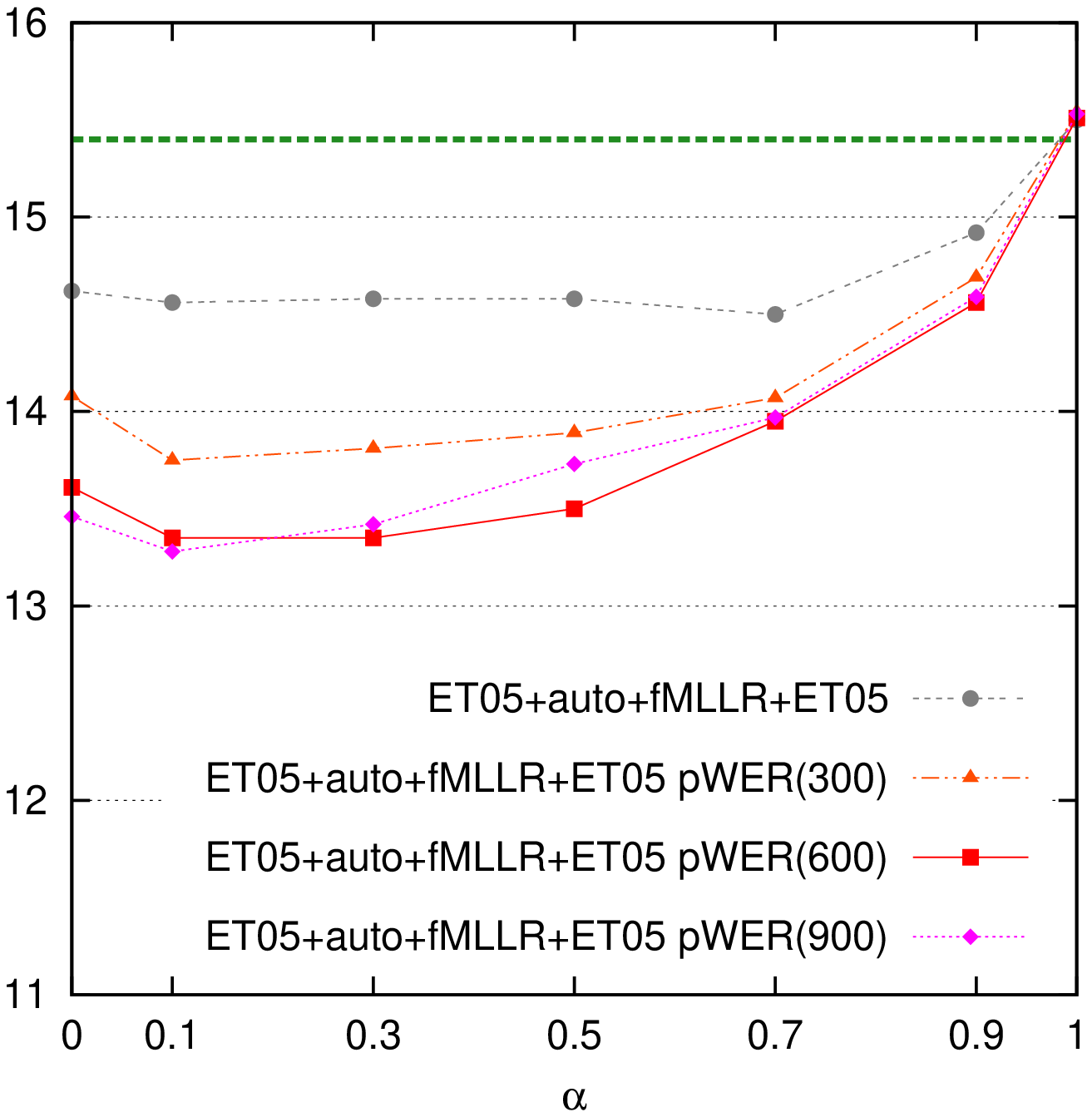}}
\caption{}
\end{subfigure}
\caption{WER achieved with oracle ({\em oWER}) and ASR QE ({\em pWER}) selection of adaptation utterances, on the evaluation set {\em ET05}, as functions of the regularization coefficient $\alpha$.}
\label{fig:et05_real_oracle-predicted_vs_adasize}
\end{figure}
Figures~\ref{fig:et05_real_oracle-predicted_vs_adasize}a and~\ref{fig:et05_real_oracle-predicted_vs_adasize}b were derived, similarly to Figures~\ref{fig:dt05_real_oracle-predicted_vs_adasize}a and~\ref{fig:dt05_real_oracle-predicted_vs_adasize}b,  from the evaluation corpus {\em ET05}.

From 
the above figures, it is evident the efficacy of using  only subsets of mid-high quality transcriptions for adapting the DNNs employed in the second decoding pass. Indeed, in each figure the minimum WER is reached with  a couple of optimal values of the pair: $(\alpha,K)$, where $K$ is the size of the adaptation set. 
This value is $900$ for {\em DT05} (Figures~\ref{fig:dt05_real_oracle-predicted_vs_adasize}a and\ref{fig:dt05_real_oracle-predicted_vs_adasize}b) and $600$ for {\em ET05} (Figures~\ref{fig:et05_real_oracle-predicted_vs_adasize}a and\ref{fig:et05_real_oracle-predicted_vs_adasize}b). 
\mn{The total improvement with respect to \textit{i)} the baseline performance and \textit{ii)} the performance achieved using the whole set of adaptation utterances is remarkable.}
The difference in the optimal values of $K$ for {\em DT05} and {\em ET05} is probably due to the different size of the two corpora ({\em DT05} contains 1,640 utterances, {\em ET05} contains 1,320 utterances). 
Unsurprisingly, the performance achieved with the ASR QE approach 
is lower than the upper-bound 
results obtained with oracle WER estimates. %
However, especially on  the evaluation corpus {\em ET05}, the improvements over the baseline  are considerable. 

Note that, in all figures, the optimal values of $\alpha$ result to be quite low, ranging in the interval $[0.1-0.3]$. This \df{indicates that the regularization term contributes to a small extent to the reduction of the overall WER. Consequently,  the performance improvements resulting after utterance selection can be mostly attributed to  the better quality of the adaptation set (the adaptation method trusts  adaptation sets that now include   only ``good'' sentences) rather than to the use of KLD regularization.}

Although for comparison purposes our analysis  focused on the size of the adaptation set to perform sentence selection, in real applications it is more feasible to select sentences on the basis of  their 
predicted WER. Therefore, the next set of experiments  
was carried out by putting selection thresholds on sentence  ASR QE  predictions.
\begin{figure}[!h]
\centering
\begin{subfigure}[b]{0.48\textwidth}
{\includegraphics[width=7.5cm,angle=-90]{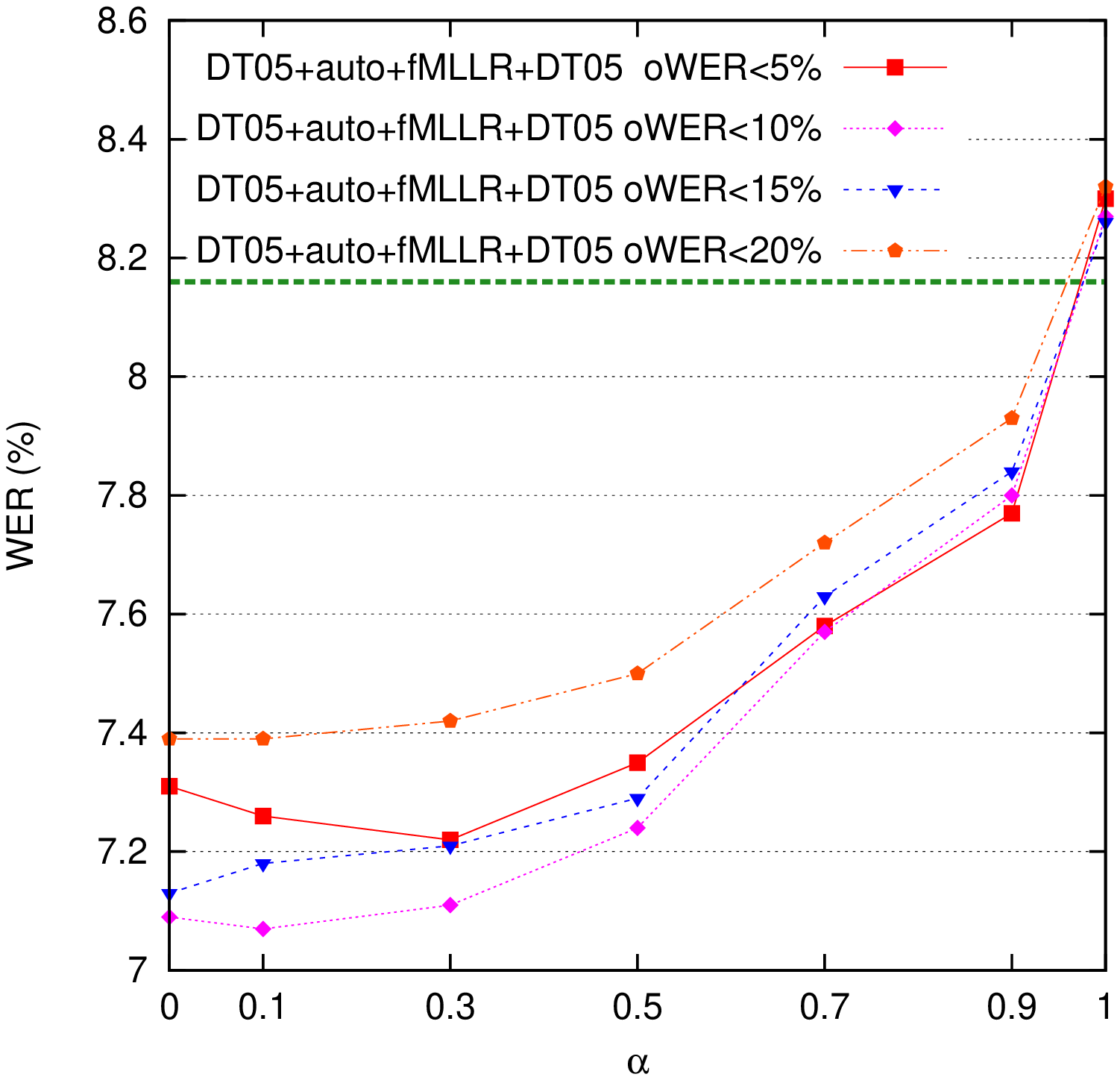}}
\caption{}
\end{subfigure}
\begin{subfigure}[b]{0.48\textwidth}
{\includegraphics[width=7.5cm,angle=-90]{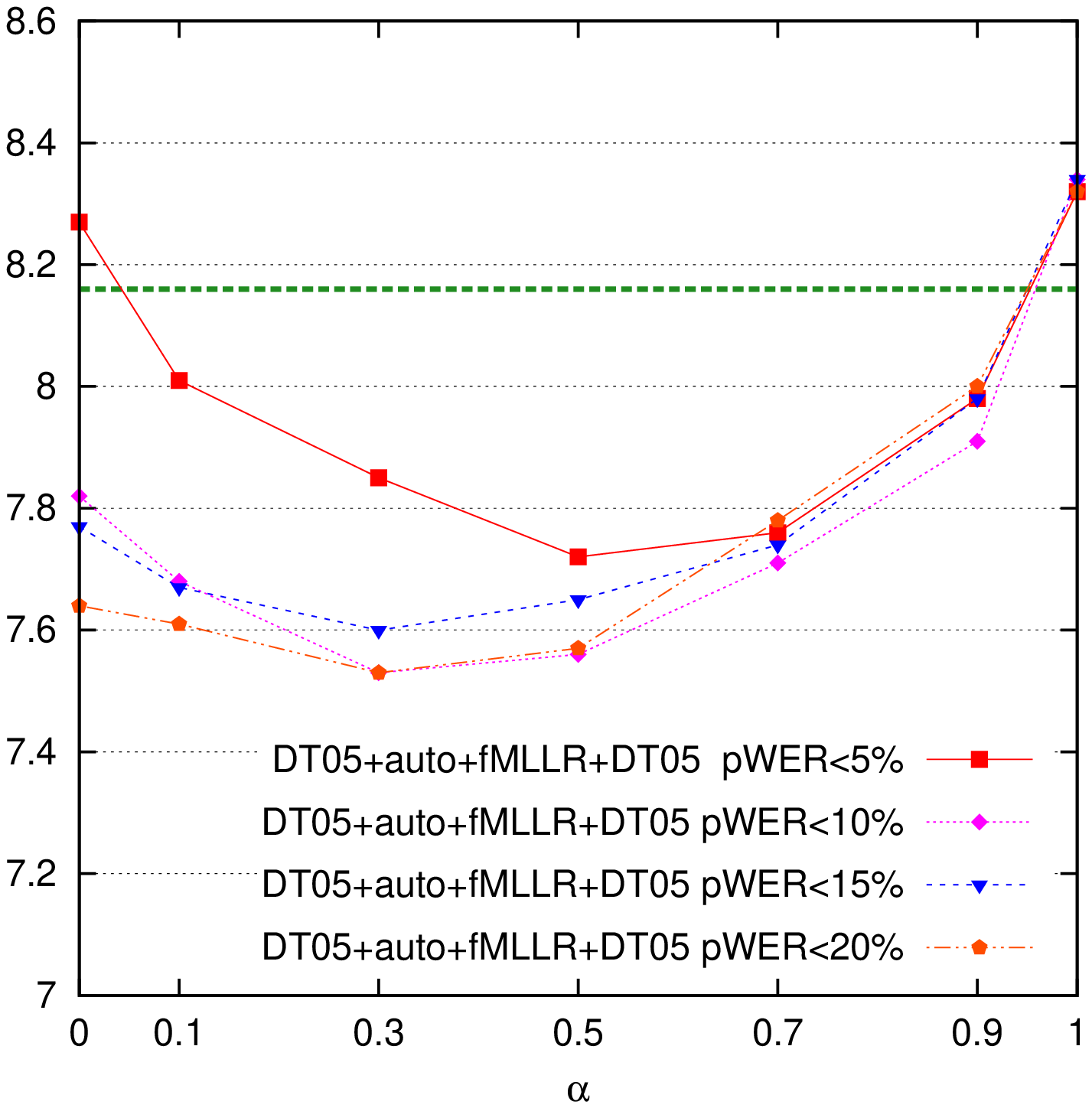}}
\caption{}
\end{subfigure}
\caption{WER achieved with oracle ({\em oWER}) and ASR QE ({\em pWER}) selection of adaptation utterances, on the development set {\em DT05}, varying the WER thresholds.}
\label{fig:dt05_real_unsup_oraclewerle_dt05_real_fmllr}
\end{figure}
Figure~\ref{fig:dt05_real_unsup_oraclewerle_dt05_real_fmllr}a shows the performance achieved on the {\em DT05} corpus with hard DNN adaptation as a function of the $\alpha$ coefficient,  varying the  thresholds applied to {\em oWER} values to select the adaptation utterances. Figure~\ref{fig:dt05_real_unsup_oraclewerle_dt05_real_fmllr}b, instead, shows the 
performance reached when {\em pWER} estimates are employed. 
Also in this case, the performance improvements  with respect to the baseline,  with both {\em oWER} and {\em pWER} 
are evident.
The optimal values for  the pair {\em oWER}$_{thr}$, $\alpha$ (Figure~\ref{fig:dt05_real_unsup_oraclewerle_dt05_real_fmllr}a) resulted to be
{\em oWER}$_{thr}=10\%$, $\alpha=0.1$, where {\em oWER}$_{thr}$ indicates the selection threshold. Similarly, using {\em pWER} estimates the corresponding optimal values are {\em pWER}$_{thr}=10\%$, $\alpha=0.3$. 
With {\em pWER}, the higher value for $\alpha$ compared to the value  resulting from the use of the {\em oWER} ($\alpha=0.1$) approach 
is probably due to errors in the  automatic WER predictions that have to be compensated.

Table~\ref{table:summary_et05_real} gives the final performance achieved on both {\em DT05} and {\em ET05}   using the two-pass decoding approach and the automatic selection of adaptation utterances by means of automatically predicted WERs.   For both sets, the optimal values of the pairs ($\alpha$, {\em WER}$_{thr}$) are those estimated on the {\em DT05} development corpus, i.e. ($\alpha=0.1$, {\em oWER}$_{thr}=10\%$) and  ($\alpha=0.3$, {\em pWER}$_{thr}=10\%$).

\begin{table}[!h]
\centering
\begin{tabular}{l|c|c||c|c|}
\cline{2-5}
                               & \multicolumn{2}{c||}{DT05} & \multicolumn{2}{c|}{ET05} \\ \cline{2-5} 
                               & fMLLR       & fbank       & fMLLR       & fbank       \\ \hline
\multicolumn{1}{|l|}{baseline} & 8.2         & 11.1        & 15.4        & 20.9        \\ \hline
\multicolumn{1}{|l|}{oWER}     & 7.1         & 7.7         & 12.4        & 14.2        \\ \hline
\multicolumn{1}{|l|}{pWER}     & 7.5         & 8.3         & 13.6        & 15.0        \\ \hline
\end{tabular}
\caption{\%WER achieved in homogeneous conditions using the optimal parameter pairs ($\alpha$,{\em WER}$_{thr}$) estimated on {\em DT05}.}
\label{table:summary_et05_real}\end{table}
Table~\ref{table:summary_et05_real}  
confirms the effectiveness of the proposed two-pass, \df{QE-based} adaptation approach \df{when QE parameter optimization is carried out on the development set}. In the table it  can be noticed that, although 
 filter-bank features exhibit higher WER than the fMLLR normalized ones, after the unsupervised adaptation procedure the performance gap is significantly reduced (less than 2\% absolute WER on {\em ET05}).  In all cases, the small differences between the performance yielded by the use of oracle and the corresponding predicted WERs is noteworthy.

\df{The  results in Table~\ref{table:summary_et05_real}  are noticeable, considering that they outperform those given by a strong ASR baseline, implemented} with 
state-of-the-art ASR technologies, i.e.: BeamformIt for speech enhancement, hybrid DNN-HMMs for acoustic modeling and speaker-dependent fMLLR transformations for acoustic model adaptation. 

\subsubsection{LM rescoring}

Table~\ref{table:summary_et05_real_rescoring} illustrates the results achieved on {\em ET05} with the LM  rescoring procedure released in the updated CHiME-3 recipe.
This procedure rescores \df{the final word lattices produced in the second decoding pass   by two consecutive steps: first by using a  5-grams LM, then by means of a linear combination of  a 5-grams LM and a RNNLM.}

\begin{table}[!h]
\begin{center}
\begin{tabular}{|l|c|c|c|} 
\hline
&  3-gram  & 5-gram & RNNLM \\\hline \hline
oWER & 12.4 & 10.8 & 9.9  \\ \hline
pWER & 13.6 & 11.9 & 10.9  \\ \hline
\end{tabular}
\caption{\%WER achieved, in homogeneous conditions   on {\em ET05},  with automatic data selection and using the baseline LM rescoring passes (see~\cite{hori2015}). }
\label{table:summary_et05_real_rescoring}
\end{center}
\end{table}

The 
significant performance gains demonstrate the additive effect of  LM rescoring over DNN adaptation, allowing us to reach a significant $10.9$\% WER on {\em ET05}.\footnote{See \url{http://spandh.dcs.shef.ac.uk/chime_challenge/results.html} for the official results of the challenge.}

\section{Discussion}
\label{sec:res_fmllr}


The  results achieved so far allow us to claim that, regardless of the type of acoustic features employed in the experiments (filter-bank or fMLLR normalized):
\begin{enumerate}
    \item[{\em a})] The benefits yielded by  KLD-based regularization, compared with  DNN retraining without any regularization, are limited. This is probably due to the fact that  the size of the adaptation sets considered in our experiments  is large enough to prevent data overfitting (actually,  
    previous research on KLD regularization \cite{yu2013} demonstrates its effectiveness using only few minutes of adaptation data);
    \item[{\em b})]  The presence of errors in the automatic transcription of the adaptation data is detrimental, especially when DNN adaptation is carried out in  homogeneous conditions. In fact, comparing the results in the last two rows of Table~\ref{table:expeunsup} (achieved by using the whole {\em ET05} corpus as adaptation set) with those in Table~\ref{table:summary_et05_real} (obtained by using a subset of adaptation utterances with ``few'' transcription errors) we notice, in oracle conditions, absolute WER reductions of around 2\% with fMLLR  and 4\%  with filter-bank features. Coherent WER reductions of around 1\% and 3\% are also achieved when applying our ASR QE-based selection method. This demonstrates  the effectiveness of the proposed QE-informed  approach for DNN adaptation.
\end{enumerate}  

It is worth pointing out that, till now, we have only considered KLD regularization for implementing DNN adaptation.  However, as mentioned in Section~\ref{sec:related}, several previous works proposed alternative approaches based on the use of a single linear transformation, which can be applied either to the input or the output layer of the network.  Therefore, in order to assess the effectiveness and the general applicability of the proposed  QE-based approach, we also experimented with the output-feature discriminative linear regression (oDLR) transformation,
in a way similar to that described in \cite{yao2012}. 
The results obtained in homogeneous conditions, both with and without ASR QE, are given in Table~\ref{table:oDLR_et05_real} (for comparison purposes, the baseline performance is also reported in the table). Similarly to results shown in Table~\ref{table:summary_et05_real}, the optimal thresholds for both oracle and predicted sentence WER values are empirically estimated on {\em DT05}. The resulting values for $oWER_{thr}$ and $pWER_{thr}$ are respectively 10\% and 20\%.



\begin{table}[!h]
\centering
\begin{tabular}{l|c|c||c|c|}
\cline{2-5}
                               & \multicolumn{2}{c||}{DT05} & \multicolumn{2}{c|}{ET05} \\ \cline{2-5} 
                               & fMLLR       & fbank      & fMLLR    & fbank       \\ \hline
\multicolumn{1}{|l|}{baseline} & 8.2         & 11.1       & 15.4     & 20.9     \\ \hline
\multicolumn{1}{|l|}{oDLR}     & 7.9         & 9.6        & 13.8     & 17.5        \\ \hline
\multicolumn{1}{|l|}{oDLR+oWER}& 7.4         & 9.2        & 13.0     & 16.8        \\ \hline
\multicolumn{1}{|l|}{oDLR+pWER}& 7.7         & 9.5        & 13.6     & 17.2      \\ \hline
\end{tabular}
\caption{\%WER achieved in homogeneous conditions with oDLR-based adaptation, without using ASR QE (oDLR), using utterance selection based on oracle WERs (oDLR+oWER) and on predicted WERs (oDLR+pWER). 
}
\label{table:oDLR_et05_real}
\end{table}

As 
shown in the table, 
the use of oDLR alone (even without ASR QE) always results in noticeable improvements over the baseline. 
The considerable WER reductions measured in oracle conditions (oDLR+oWER), however, indicate the high potential of a QE-driven  selection of the adaptation utterances also with this
simple 
DNN adaptation method. In general, the performance improvements are smaller than the corresponding results for KLD regularization  reported in Table~\ref{table:summary_et05_real}.
Such lower 
results 
can be explained by the 
findings reported in \cite{gollan2008}, in which the authors 
compared approaches based on MLLR and maximum a posterior probability (MAP) for GMM-HMMs adaptation. In this case, the impact of errors in the supervision 
is directly  proportional to the number of transformation parameters to estimate. Indeed, while in the experiments reported in Section~\ref{sec:results} all the parameters of the original DNN are adapted,  with oDLR  only a small fraction of them (around 13\%) is updated.
The reduced sensitivity to  errors in the supervision is also reflected by the higher value of the threshold  used to select the adaptation data (20\% for oDLR vs 10\% for KLD).

The results measured in oracle conditions suggest 
a higher potential for the application of QE to KLD-based regularization  rather than to oDLR. This intuition, however, is partially contradicted by the last row of Table~\ref{table:oDLR_et05_real} (oDLR+pWER). 
With predicted WER scores, indeed, 
the values achieved with fMLLR are  only slightly worse or identical
to those in  Table~\ref{table:summary_et05_real}.
%
To put into perspective this unexpected ``exception'' in the results, it's worth remarking that the impact of QE in DNN adaptation is proportional to the acoustic mismatch between  training and  test data. As observed in Sections~\ref{subsubsec:all1} and \ref{subsubsec:all2}, fMLLR features have the capability to reduce such mismatch, making the gains brought by QE-based adaptation less evident than those achieved with filter-bank. In light of this, although on \textit{ET05} and with fMLLR features oDLR is competitive with the more complex KLD-based regularization  proposed in this paper, we believe that more challenging data (featuring a higher mismatch between training and  test) would increase the distance between the two approaches and reward our method.
A comparison between different DNN adaptation methods across multiple data sets featuring variable degrees of acoustic mismatch is definitely an interesting direction for future research.

\section{Conclusions}
\label{sec:conclusions}

\mn{In this paper, we proposed to exploit  automatic Quality Estimation (QE) of ASR hypotheses to perform the unsupervised adaptation of a deep neural network modeling acoustic
probabilities.}
We developed our approach motivated by the two following hypotheses:
\begin{enumerate}
    \item The adaptation process does not necessarily require the supervision of a manually-transcribed development set. Manual supervision can be replaced by a two-pass decoding procedure, in which  the evaluation data we are currently trying to recognise are automatically transcribed and used to inform the adaptation process;
    \item The whole process can benefit from methods that take into account the quality of the supervision. In particular, automatic quality predictions can be used either to weigh the adaptation instances or to discard the less reliable ones.
\end{enumerate}

To implement our approach, we retrained a (baseline) DNN  by minimizing an objective function defined by a linear combination  of the usual cross-entropy measure (evaluated on a given adaptation set) and a regularization term. This is the Kullback-Leibler divergence  between the output distribution of the original DNN and the actual output distribution.

First, we experimented in ``cross conditions'',
by adapting on the \textit{real} development set 
of the CHiME-3 
challenge and testing on the corresponding \textit{real} evaluation set. 
In this scenario, we found that, when using all the manually-transcribed adaptation data, the KLD-based approach is effective.
Then, moving to the automatically-generated supervision of the  adaptation data, we discovered a correlation between performance results and the quality of the adaptation data. In particular, in ``oracle'' conditions (i.e. with true WER scores), DNN adaptation benefits from removing utterances with a WER score  above a given threshold.

%

Building on this result,  we focused on ``self'' DNN adaptation in ``homogeneous conditions'',
in which the baseline DNN is adapted on the same evaluation set ({\em ET05}) by exploiting the automatic supervision  derived from a first ASR decoding pass. \df{Similarly to the cross-condition scenario, this approach allowed us to significantly improve the performance when ``low quality'' sentences (i.e. sentences that exhibit oracle WERs higher than  an optimal threshold) are removed from the adaptation set.}
Improvements were measured not only in ``oracle'' conditions (i.e. with true WER scores), but also in realistic conditions 
in which manual references are not available and the only viable solution is to rely upon predicted WERs. 
%
To this aim, building on previous positive results on quality estimation for ASR  \cite{Negri:2014,cdesouza-EtAl:2015:NAACL-HLT,sjalalvand-EtAl:2015:ACL}, we used automatic WER  prediction as a criterion to 
isolate 
subsets of the adaptation data featuring variable quality.
%
%
The 
results of an extensive set of experiments allowed us to conclude that:


\begin{itemize}
\item \df{Exploiting ASR QE for DNN adaptation in a two-pass decoding architecture yields significant  performance improvements over the strong, most recent CHiME-3 baseline;}
\item \df{Self DNN adaptation is more effective with filter-bank acoustic features than with fMLLR normalized features. This behavior is probably due to the smaller mismatch between training and test data caused by the use of fMLLR transformations, indicating a higher potential of the QE-driven approach in a scenario characterized by weakness of  fMLLR  in reducing such mismatch (e.g. with small adaptation sets);} 
\item \df{ASR QE is less effective with output discriminative linear regression (oDLR) transformation for DNN adaptation,  due to the lower  number of parameters to adapt compared to KLD regularization. 
This demonstrates the portability of our method, but a  higher effectiveness  with  large DNNs.}

\end{itemize}

Finally, we applied  the LM rescoring procedure delivered with the CHiME-3 baseline to the word lattices produced after the second, DNN-adapted, decoding pass. The resulting WER reductions demonstrate the independent effects of LM rescoring and the proposed DNN adaptation approach.
Our full-fledged system for DNN adaptation, integrating KLD and ASR QE for   data selection, allows us to outperform the strong CHiME-3 baseline with a 1.7\% WER reduction (from 12.6\% to 10.9\%).


Some interesting directions for future work already emerged in the course of this research. One is to further explore the portability of the proposed ASR QE approach, by integrating it into other state-of-the-art ASR systems. To validate our working hypotheses, in this paper we started from the strong CHiME-3 baseline; now  it would  be interesting to test its effectiveness within more 
powerful DNNs (e.g.  capable of modeling time dependencies among acoustic observations, such as bidirectional recurrent neural networks \cite{graves2014,amodei2016}) having a higher number of parameters to adapt.
Another interesting direction is to investigate ways to express hypotheses' quality at a granularity level higher than that of the sentence, e.g. at the level of words or even  single frames. 
To do this we also  plan to replace  the ``\textit{ad hoc}'' formula expressed by Equation~\ref{eq:alfasent}, to weigh the KLD regularization term in Equation~\ref{eq:div1}, by jointly optimizing both the DNN weights and the sentence-dependent  regularization coefficients.

Based on the successful results obtained in 
the specific CHiME-3 application framework (read speech acquired by multiple microphones in noisy conditions), we also plan to extend our approach to other domains, 
possibly featuring higher degrees of acoustic mismatch for which  the KLD-based regularization proposed
in this paper seems to have the highest potential. 
%
%
%
%
\df{Finally, an interesting  direction to investigate is ``incremental'' DNN adaptation, where the DNN is periodically adapted  on speech utterances and related transcriptions  stored after they have been processed  by the ASR system. This application scenario reflects the cross-condition experimental situation defined in Section~\ref{ssec:expe}
and, based on our current results, represents a promising and natural extension of this research.}

 \bibliographystyle{model2-names}

\bibliography{sample2}

\end{document}